\documentclass{article}

\usepackage{arxiv}
\usepackage{amsfonts}
\usepackage{url}
\usepackage{booktabs}
\usepackage{subfig}
\usepackage{placeins}
\usepackage{amsmath}
\usepackage[capitalise,noabbrev]{cleveref}
\usepackage{graphicx}
\usepackage{multicol,multirow}

\newcommand\add[1]{\textcolor{black}{#1}}
\usepackage{enumitem}
\usepackage{xcolor}
\newcommand{\comm}[2]{{\textcolor{#1}{#2}}}

\newcommand{\gm}[1]{\comm{black}{#1}}

\usepackage[normalem]{ulem}
\usepackage{siunitx}
\usepackage{xspace}
\makeatletter
\DeclareRobustCommand\onedot{\futurelet\@let@token\@onedot}
\def\@onedot{\ifx\@let@token.\else.\null\fi\xspace}
\DeclareRobustCommand\nodot{\futurelet\@let@token\@nodot}
\def\@nodot{\ifx\@let@token.\else~\null\fi\xspace}

\def\ie{\emph{i.e}\onedot}

\def\wrt{w.r.t\onedot}

\def\etal{\emph{et al}\onedot}
\makeatother

\begin{document}

\title{Sketch and Patch: Efficient 3D Gaussian Representation for Man-Made Scenes}

\author{
  \textbf{Yuang Shi}$^{1}$,
  \textbf{Simone Gasparini}$^{2}$,
  \textbf{Géraldine Morin}$^{2}$,
  \textbf{Chenggang Yang}$^{1}$,
  \textbf{Wei Tsang Ooi}$^{1}$ \vspace{1mm} \\
  $^1$National University of Singapore\quad
  $^2$IRIT - University of Toulouse \quad   \\
  \small{\texttt{\{yuangshi, chenggan, ooiwt\}@comp.nus.edu.sg}} \\
  \small{\texttt{\{simone.gasparini, geraldine.morin\}@toulouse-inp.fr}} 
  \vspace{-5mm}
}


\maketitle

\begin{figure*}[ht]
    \centering
    \includegraphics[width=\textwidth]{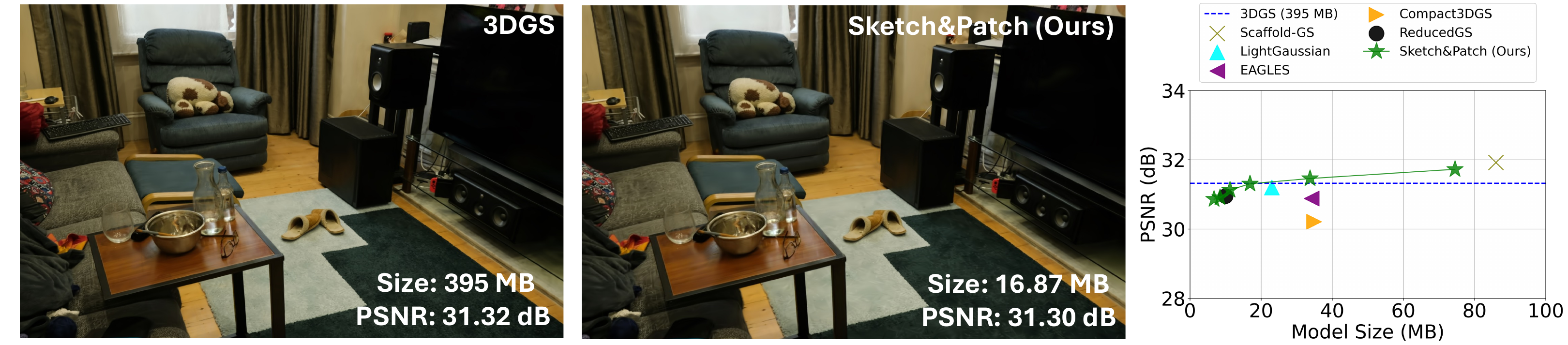}
    \caption{We propose a hybrid Gaussian representation with 3D structure prior, significantly reducing the storage of 3DGS model by an order of magnitude while maintaining the visual quality.}
\end{figure*}

\begin{abstract}
3D Gaussian Splatting (3DGS) has emerged as a promising representation for photorealistic rendering of 3D scenes. 
However, its high storage requirements pose significant challenges for practical applications. 
We observe that Gaussians exhibit distinct roles and characteristics that are analogous to traditional artistic techniques --- Like how artists first sketch outlines before filling in broader areas with color, some Gaussians capture high-frequency features like edges and contours; While other Gaussians represent broader, smoother regions, that are analogous to broader brush strokes that add volume and depth to a painting.

Based on this observation, we propose a novel hybrid representation that categorizes Gaussians into (i) \textit{Sketch Gaussians}, which define scene boundaries, and (ii) \textit{Patch Gaussians}, which cover smooth regions. 
Sketch Gaussians are efficiently encoded using parametric models, leveraging their geometric coherence, while Patch Gaussians undergo optimized pruning, retraining, and vector quantization to maintain volumetric consistency and storage efficiency. 
Our comprehensive evaluation across diverse indoor and outdoor scenes demonstrates that this structure-aware approach achieves up to \SI{32.62}{\percent} improvement in PSNR, \SI{19.12}{\percent} in SSIM, and \SI{45.41}{\percent} in LPIPS at equivalent model sizes, and correspondingly, for an indoor scene, our model maintains the visual quality with \SI{2.3}{\percent} of the original model size.
\end{abstract}

\begin{figure*}[t!]
    \centering
    \subfloat[3DGS.]{%
        \includegraphics[width=0.4\textwidth]{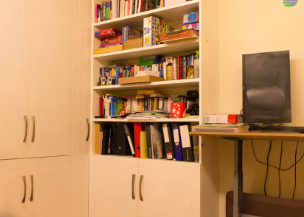}
        \label{subfig:render_image}
        }
    \subfloat[Ellipsoids.]{%
        \includegraphics[width=0.4\textwidth]{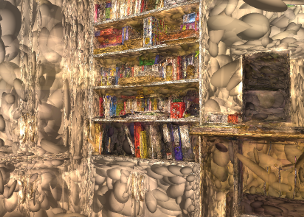}
        \label{subfig:gaussian}
        } \\
    \subfloat[Ellipsoid centers.]{%
        \includegraphics[width=0.4\textwidth]{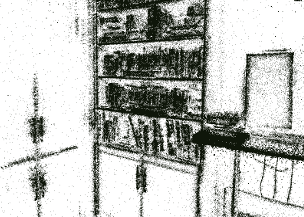}
        \label{subfig:points}
        }
    \subfloat[3D lines.]{%
        \includegraphics[width=0.4\textwidth]{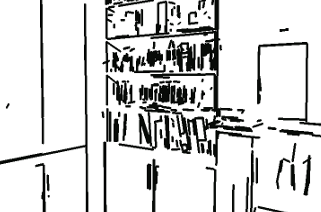}
        \label{subfig:line}
        }
    \caption{Illustration of the characteristics of 3DGS. From left to right: rendered images from (a) 3DGS model, (b) ellipsoids of 3DGS, (c) ellipsoid centers, (d) 3D lines within the model extracted by Line3D++~\cite{line3dcode}. Note the correlation between the density of 3DGS (or ellipsoid centers) and the location of 3D lines. }
    \label{fig:point_gaussian}
\end{figure*}

\section{Introduction}
\label{sec:intro}

The increasing demand for immersive experiences in extended reality (XR) applications — such as virtual reality (VR), augmented reality (AR), and cloud gaming — has driven significant progress in 3D scene representation technologies.
These applications rely on accurately modeling complex 3D environments to deliver high-quality user experiences. 

Classical 3D representations explicitly model the geometry.  Continuous models, such as parametric surfaces or meshes, support the appearance to be mapped onto them, whereas discrete models, such as point clouds, have appearance attributes such as colors, normals, and reflectance. 
Continuous models are preferred for design, but point clouds, as the typical output of scanning devices, have been used for digital twins.
However, a scan of large environments can easily produce millions of points, each with rich attributes such as colors, normals, and reflectance, posing significant storage challenges. 
Additionally, the discrete nature of point clouds can easily introduce visual artifacts~\cite{shi2024volumetric} when compressed. 
Furthermore, when point clouds are projected to 2D for rendering, they can have holes~\cite{kerbl20233D} that disrupt the visual experience and hinder the seamless immersion of the viewers.

Recently, learning-based methods have given rise to alternative representations that seamlessly model geometry and appearance. 
First, NeRF (Neural Radiance Field)~\cite{mildenhall2021nerf} (and its more than $9000$ follow-up papers) implicitly models the 3D as a multi-layer perceptron, offering the ability to render and navigate into a seamless and high visual quality virtual, reconstructed scene. 
The main drawbacks of this representation are the resources required to train the model from a large number of images and the performance for inferring new viewpoints.
Alternatively 3D Gaussian Splatting (3DGS) \cite{kerbl20233D} has emerged as a leading approach for learning more explicit and more efficient point-based 3D representations while keeping the mix between geometry and appearance, leading to a very satisfying visual quality.
Compared to point clouds, 3DGS represents 3D scenes with a collection of 3D ellipsoidic Gaussian splats (or just Gaussians for short), which can mitigate the presence of holes with much sparser distribution. 
By leveraging the properties of Gaussians and taking advantages of modern GPU, 3DGS has demonstrated its ability to achieve photorealistic rendering quality and real-time rendering speed, showing great promise as a foundational representation for future immersive multimedia systems and applications. 

While 3DGS achieves impressive visual fidelity, it still generates vast numbers of Gaussians to capture fine geometric and appearance details, with five attributes stored independently. 
For instance, Kerbl~\etal~\cite{kerbl20233D} employ one to five million Gaussians to model static scenes, demanding up to \SI{1}{\giga\byte} of storage per scene, which impedes efficient transmission and rendering.

Common point-based streaming strategies, such as 3D downsampling or tiling, are designed to address storage issues in 3D point clouds by generating scalable and multi-resolution representations~\cite{han2020vivo,viola2023volumetric,shi2023enabling,shi2024qv4}. 
However, these methods are unsuitable for 3DGS due to the unique properties of its Gaussian-based representation. 
Unlike conventional point clouds, 3DGS employs heterogeneous Gaussians parameterized by ellipsoid shape and view-dependent color. 
These attributes are optimized iteratively to minimize discrepancies between rendered and ground-truth images, leading to a nonuniform and highly interdependent distribution of Gaussians, as shown in \cref{subfig:gaussian}. 
This tightly coupled representation optimizes a set of Gaussians, thus significantly limiting the effectiveness of 3D downsampling or partitioning techniques, which often result in substantial degradation of the visual quality.

Additionally, 3DGS's adaptive density control mechanism introduces storage inefficiencies.
The adaptive density control aims to enhance fidelity by cloning Gaussians in under-reconstructed regions and splitting those in over-reconstructed areas, based on 2D positional gradients. 
The reliance on 2D gradients often overestimates the need for densification in high-frequency and boundary-defining areas, such as edges and contours. 
For instance, \cref{subfig:points} depicts the distribution of ellipsoid centers in the scene. As illustrated, this process results in dense Gaussian clusters along structural boundaries, like architectural edges, furniture boundaries, and other manufactured object contours, leading to redundant representation and increased computational overhead~\cite{kim2024color,li2025geogaussian}.

Based on these characteristics of 3DGS, we propose to categorize Gaussians into two distinct roles: \textit{Sketch Gaussians} and \textit{Patch Gaussians}, drawing inspiration from traditional artistic techniques.
Like how artists first sketch outlines before filling in broader areas with color, Sketch Gaussians capture boundary-defining features such as edges and contours, serving as the semantic scaffolding of the scene. These Sketch Gaussians exhibit strong coherence along linear structures, particularly in man-made scenes where edges and contours often follow consistent geometric patterns.
In contrast, Patch Gaussians, analogous to broader brush strokes that add volume and depth to a painting, provide volumetric coverage for smoother and broader regions.

We further propose leveraging the unique structural properties of edges and contours, which typically represent high-frequency and boundary-defining features that are inherently linear or curvilinear. 
Specifically, instead of densely populating these areas with Sketch Gaussians explicitly, we propose encoding Sketch Gaussians with the prior of 3D line segments. 
By abstracting the coherent Sketch Gaussians into compact parametric \gm{models}, such as polynomials or splines, we preserve both the visual quality and semantic information with data of significantly smaller size. 
For smoother and low-frequency regions, the Patch Gaussians are stored with their Gaussian parameters, given the naturally sparse distribution of Gaussians in these areas.
This dual-role categorization forms the basis of our hybrid Gaussian representation, designed to enhance scalability, structure awareness, and storage efficiency for 3DGS.

\begin{figure*}[t]%
    \centering
    \includegraphics[width=\textwidth]{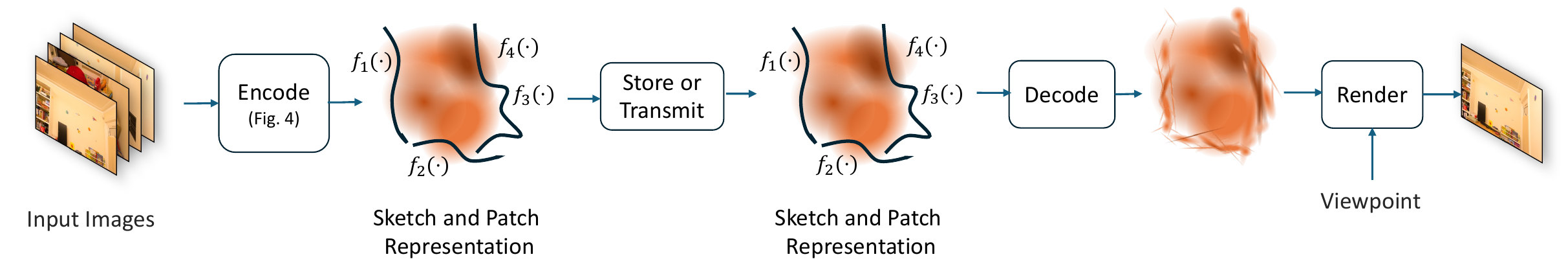}%
    \caption[]{The streaming pipeline. We propose to build an efficient representation by leveraging the 3D structure information, which can be essentially regarded as a ``codec'' in the context of the streaming system.}%
\label{fig:pipeline}
\end{figure*}

While our approach works for any 3DGS scenes, it is especially effective in man-made (as opposed to natural) scenes and objects, with well-defined straight edges and planar surfaces that naturally align with our Sketch and Patch categorization.  
The prevalence of regular geometric primitives in architectural elements, furniture, and manufactured objects enables the effective separation of boundary-defining features from broader regions, allowing for efficient encoding through our hybrid representation strategy (see \cref{fig:pipeline}).

Our contributions can be summarized as follows.
\begin{itemize}
    \item A novel dual-role categorization of 3DGS into Sketch and Patch Gaussians, reflecting their distinct functions in scene representation.
    \item A hybrid Gaussian representation that encodes high-frequency and boundary-defining features with structural 3D line prior 
    and preserves volumetric detail for low-frequency regions using 3D Gaussians.
    \item Relying on the intrinsic structure of man-made scenes, our approach achieves consistent improvements in visual quality by up to \SI{32.62}{\percent} in PSNR, \SI{19.12}{\percent} in SSIM, and \SI{45.41}{\percent} in LPIPS at the equivalent model size and correspondingly, a proposed hybrid model of size around \SI{2}{\percent} of the original model can maintain the visual quality of an indoor scene.
\end{itemize}

The rest of this paper is organized as follows. 
\cref{sec:related} presents the background of 3DGS and related works on 3D scene abstract representation and compact 3DGS representation. 
\cref{sec:method} introduces the framework we proposed to construct the efficient hybrid Gaussian representation. 
Experiments and results are presented in \cref{sec:experiments}. 
Finally, we discuss and conclude this paper in \cref{sec:conclude}.

\section{Background and Related Work}
\label{sec:related}

\subsection{3D Gaussian Splatting}
\label{subsec:3D_gaussian_splatting}

3D Gaussian splatting (3DGS)~\cite{kerbl20233D} proposes a 3D reconstructed model capable of novel view synthesis, known for its high reconstruction quality, faster training speed, and real-time rendering through rapid rasterization. 
The 3DGS represents a scene using a collection of Gaussian Splats optimized to fit a set of input images, compared with a splatting-based~\cite{zwicker2001ewa} rendering of the model.

Each Gaussian is parameterized by a center position $\mathbf{\mu} \in \mathbb{R}^{3\times3}$, a 3D covariance matrix $\mathbf{\Sigma}$, an opacity $\alpha$, and a color $c$ which is described by a set of Spherical Harmonic (SH) coefficients $\mathbf{K}$. 
A Gaussian centered at $\mathbf{\mu} $ is defined as:
\begin{equation}
        G(x) = e^{-\frac{1}{2}(x)^T\Sigma^{-1}(x)}.
\end{equation}

To construct a collection of 3D Gaussians that accurately captures the scene’s essence, 3DGS introduces an optimization method using differentiable rendering to estimate the parameters of the 3D Gaussians, fitting a set of calibrated input images of the given scene. 
During optimization, 3DGS iteratively renders 2D images from the training views and minimizes the loss between the rendered images and the ground-truth input images. 
The loss function is a combination of the $\mathcal{L}_1$ loss and the Structural Similarity loss (SSIM)~\cite{wang2004image}, weighted by
\begin{equation}
        \mathcal{L} = \lambda \cdot \mathcal{L}_1 + (1 -\lambda) \cdot \mathcal{L}_{SSIM}.
\end{equation}

During the 3DGS training process, Gaussian points adapt to better fit their surroundings. 
Density is controlled by a positional gradient threshold, balancing between the number of parameters and the fitting accuracy. 
Starting from a sparse point set generated by Structure from Motion (SfM), the model iteratively removes the Gaussian splats whose opacities are below the pre-set threshold, and densifies the Gaussian splats in regions with high positional gradients. 
In areas lacking detail, low-density Gaussians are duplicated and shifted along the gradient to create new geometry. 
In areas of excessive overlap, large Gaussians are split into smaller ones to achieve finer granularity. 

This adaptive density control refines the scene representation while managing the number of Gaussians to balance model complexity and rendering quality.

\subsection{3D Scene Abstract Representations}
\label{subsec:3D_scene_abstract_repres}
Our work proposes to benefit from the correlation between the scene structure and the 3DGS representation (see \cref{fig:point_gaussian} $(c)$ and $(d)$). 
In the following, we review existing methods to abstract a 3D scene by analyzing the scene to identify its features, particularly linear (1D) features.
3D scene abstraction varies depending on the types of input and output. 
For instance, input types can include 2D images or 3D data, while output types may consist of features like line segments, curves, or edges that combine both.

In earlier methods, extracting features from 2D inputs to abstract 3D line segments typically relies on Structure from Motion (SfM) algorithms~\cite{Hartley2000,schonberger2016structure, wu2013towards}. 
SfM reconstructs 3D structures from unordered image sets by simultaneously estimating intrinsic and extrinsic camera parameters and a sparse 3D point cloud. 
With SfM results, the mapping from 2D planes to 3D space becomes feasible. 
Some methods extract features from images to identify or abstract line descriptors — 2D line segments on the image plane — using classical techniques~\cite{ipol.2012.gjmr-lsd, huang2020tp, zhang2021elsd}, or neural network-based methods such as DeepLSD~\cite{pautrat2023deeplsd}. 
These 2D descriptors are then matched across multiple views to identify the same or similar line segments visible from different viewpoints. 
This process is influenced by the range of viewpoint coverage and occluded areas. 
Once matching is complete, geometric methods like triangulation~\cite{baillard1999automatic, zhang2014structure, ramalingam2015line, micusik2017structure} or other epipolar constraints~\cite{hofer2014improving, hofer2015line3d, hofer2017efficient}  are used to back project the segments into 3D space. 
Some approaches further leverage deeper geometric relationships, such as planes~\cite{wei2022elsr}, while
Schindler~\cite{schindler2006line} introduced the Manhattan-world assumption to add geometric constraints, enhancing robustness in structured environments. 
Despite significant progress in recent works~\cite{hofer2017efficient, wei2022elsr, liu20233d}, abstracting 2D data into 3D line segments still faces inherent limitations. 
These include robustness issues introduced by geometric computation errors, which can affect the final results. 
Similarly, generating curves~\cite{schmid2000geometry, kahl2003multiview} faces similar limitations.

Scene abstraction can also be achieved directly from 3D point clouds or other 3D data structures. 
Such methods directly classify and identify edge regions within 3D point clouds, which are then abstracted or parameterized into line segments or curves. 
Processing directly on 3D data avoids occlusion issues caused by multi-view matching but faces challenges from the inherent noise in 3D data and the complexities of parameter tuning during abstraction.
Some works~\cite{schnabel2007efficient, chen2017fast, wang2020pie} have addressed these challenges by including point filtering steps, weight loss functions, and other robust parametric estimations.
Also, recent methods~\cite{ye2023nef, li20243d, xue2024neat} extract 3D data containing only edge regions from 2D images to reduce noise inherent to reconstructed 3D data.
Methods~\cite{liu2021pc2wf, ma20223d} leverage deep neural networks trained on large annotated datasets to learn hidden edge features within point cloud distributions, ultimately predicting a 3D wireframe.

\subsection{Compact 3DGS Representation}

While 3DGS~\cite{kerbl20233D} leverages efficient techniques such as anisotropic Gaussians~\cite{zwicker2001ewa}, tile-based sorting, and approximate $\alpha$-blending to achieve high-performance rendering, the presence of redundancy in the set of  Gaussian splats continues to impact computational efficiency. 
Minimizing these redundancies is crucial for optimizing model performance across various applications.

The model size in 3DGS is primarily affected by two factors. 
First, complex, high-frequency areas require sophisticated parameters to represent. In particular, advanced spherical harmonics coefficients are needed for accurate modeling. 
Second, as discussed in~\cref{subsec:3D_gaussian_splatting}, the model generation process introduces additional Gaussian splats based on density thresholds during scene fitting. The combination of these factors --- the parameter complexity per Gaussian and the total number of Gaussian splats --- leads to significant storage overhead.

Recent research has approached these challenges from multiple angles. To address parameter complexity, several methods have been proposed: region-based vector quantization~\cite{niedermayr2024compressed}, K-means codebooks~\cite{navaneet2023compact3d}, and view-direction exclusion~\cite{cai2025radiative}. Fan \etal~\cite{fan2023lightgaussian} employ knowledge distillation~\cite{hinton2015distilling} to compress spherical harmonics parameters, while Lee \etal~\cite{lee2024compact} utilize learned binary masks and grid-based neural networks as alternatives to spherical harmonics.

Parallel efforts have focused on managing Gaussian density through various Level of Detail (LOD) techniques~\cite{ren2024octree,lu2024scaffold,kerbl2024hierarchical,shi2025lapisgs}. 
For instance, Ren \etal~\cite{ren2024octree} propose an octree-based organization where each level corresponds to anchor Gaussian splats defining different LODs. 
Their approach, enhanced by Scaffold-GS~\cite{lu2024scaffold}, combines anchor Gaussians with MLPs for anchor-level feature estimation, resulting in improved representation efficiency. 
Shi \etal introduce LapisGS~\cite{shi2025lapisgs}, a layered representation scheme enabling progressive and continuous quality adaptation for bandwidth-aware streaming applications.

These methods, however, primarily focus on either parameter compression or LOD structuring without considering the inherent roles of different Gaussian splats in scene representation. 
Our approach fundamentally differs by recognizing and leveraging the distinct characteristics of Gaussian splats in man-made scenes, categorizing them into Sketch and Patch components based on their geometric significance. 
This structure-aware strategy enables more efficient representation by applying appropriate encoding techniques to each category: compact parametric models for boundary-defining features and optimized pruning for volumetric regions.

Meanwhile, our Sketch and Patch categorization is complementary to these existing compression and LOD techniques. 
Parameter compression methods like vector quantization~\cite{niedermayr2024compressed}, codebook learning~\cite{navaneet2023compact3d}, or SH compression~\cite{fan2023lightgaussian,lee2024compact} can be applied to both Sketch and Patch Gaussians while maintaining their distinct roles. 
This compatibility ensures that our method can serve as a foundation for further optimizations while providing its unique benefits in storage efficiency and visual quality preservation.

\begin{figure*}[t!]%
    \centering
    \includegraphics[width=\textwidth]{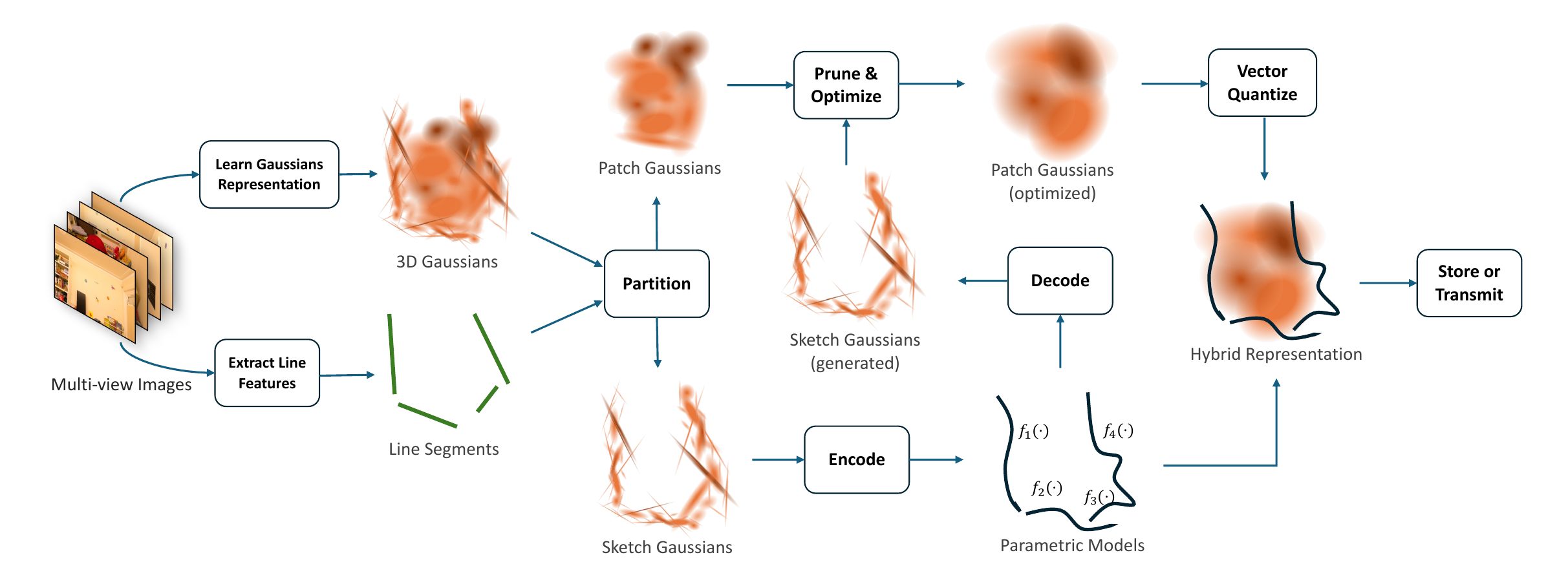}%
    \caption[]{The encoding into our compact hybrid representation. Our method involves extracting 3D line segments to abstract edges and contours, identifying Sketch Gaussians, and encoding them with compact parametric models. 
    For smoother regions, Patch Gaussians are pruned and retrained together with the decoded Sketch Gaussians to ensure sparsity while preserving visual fidelity. Vector quantization is then applied to further reduce the storage of the optimized Patch Gaussians.
    The resulting hybrid representation combines Sketch and Patch Gaussians for efficient storage, transmission, and rendering of 3D scenes.}%
\label{fig:methodology}
\end{figure*}

\section{Proposed Methodology}
\label{sec:method}

While 3DGS can represent complex 3D scenes, the globally estimated set of Gaussian splats is non-uniform, due to heterogeneous attributes; also, the adaptive density control mechanism creates numerous Gaussian splats, which leads to storage inefficiencies and thus challenges the need for more scalable and efficient representations.

To address these challenges, we propose a hybrid Gaussian representation tailored for man-made scenes. 
Our method categorizes Gaussians into Sketch Gaussians and Patch Gaussians. 
Sketch Gaussians capture high-frequency, boundary-defining features and are compactly encoded using 3D line segments and parametric models. 
In contrast, Patch Gaussians represent low-frequency, smoother regions and are optimized for storage efficiency through retraining and quantization. 
This categorization allows us to leverage the structural properties of the scene and its correlation to 3DGS density and bias (see \cref{fig:point_gaussian}), achieving a representation that balances scalability, efficiency, and visual quality.

The proposed method consists of the following steps (see \cref{fig:methodology}). 
First, we extract 3D line segments that abstract edges and contours to identify Sketch Gaussians. 
Based on the prior of line segments, we identify the Sketch Gaussians using a two-step process involving radius search and RANSAC-based filtering, enhanced with scaling-awareness to improve precision for elongated Gaussians (\cref{subsec:scene_abstract}). 
The Sketch Gaussians are then encoded using parametric models on a given 3D line to model their attributes compactly, significantly reducing storage requirements. 
For smoother regions, represented by Patch Gaussians, we first apply pruning and retraining to sparsify their distribution while maintaining visual fidelity and then perform vector quantization to further reduce the model size (\cref{subsec:encode_optimize}). 
Finally, the Sketch and Patch Gaussians are integrated into a unified hybrid representation, enabling efficient storage, transmission, and rendering of 3D scenes.
\cref{fig:methodology} illustrates our framework.

\subsection{Partitioning Sketch and Patch Gaussians} 
\label{subsec:scene_abstract} 
The first step of the pipeline analyzes the 3D scene to extract linear features to partition the set of Gaussian splats into Sketch Gaussians, associated with a given set of 3D lines, and Patch Gaussians.

\textbf{Extraction of 3D Line Segments}. 
Geometry-guided line reconstruction methods extract edges, contours, or high-frequency regions corresponding to high Gaussian splats density  (see \cref{fig:point_gaussian} $(c)$ and $(d)$). 
As explained in \cref{subsec:3D_scene_abstract_repres}, 3D features are extracted from the 2D input images.
Here, we choose the line reconstruction method \textit{Line3D++}~\cite{hofer2014improving,hofer2015line3d,hofer2017efficient}. 
It takes as input the images and camera poses and provides as output the relevant 3D lines by abstracting pixel or object boundaries into discrete 2D line segments~\cite{felzenszwalb2004efficient,von2008lsd} and back-projecting them into 3D space.

\textbf{Identification and modeling of Sketch Gaussians}.
Identifying Sketch Gaussians is a critical step in constructing a structure-aware representation. 
These Gaussian splats, concentrated along edges and contours, play an important role in preserving the semantic and visual essence of the scene. 
However, their identification is nontrivial due to the densely coupled and anisotropic nature of Gaussian splats in 3DGS. 
To address this, we propose a two-step robust process for extracting Sketch Gaussians based on structural priors provided by 2D line segments: (i) radius search and (ii) random sample consensus (RANSAC) filtering~\cite{fischler1981random}.

\textit{Sketch Gaussians along a 3D line}. In the first step, we identify a rough set of Gaussian splats associated with a line segment extracted.  
Let $\mathbf{L}_i(t)$ denote a 3D line segment parameterized as: $\mathbf{L}_i(t) = (1 - t) \mathbf{p}_{\text{start}} + t \mathbf{p}_{\text{end}}, t \in [0, 1]$, where $\mathbf{p}_{\text{start}}$ and $\mathbf{p}_{\text{end}}$ are the endpoints of the segment.

We include a Gaussian splat $\mathcal{G}_j$ for line $\mathbf{L}_i$ in the candidate set $\mathcal{S}_i$, if its center $\mathbf{\mu}_j \in \mathbb{R}^3$ lies within a radius $r$ from the line:
\begin{equation}
    d(\mathbf{\mu}_j, \mathbf{L}_i) = \min_{t \in [0, 1]} \|\mathbf{\mu}_j - \mathbf{L}_i(t)\| \leq r.
\end{equation}
The three remaining steps further ensure that the Gaussian splats selected on a line also have coherent attributes.

\textit{Robust selection of Sketch Gaussians relative to the 3D line}. After the initial radius search identifies a rough set of candidate Gaussians, the next step is to refine this set using RANSAC, to ensure coherent attribute values among the Sketch Gaussians on a given line. 
RANSAC is well-suited for this task, as it robustly fits models to data with outliers. 
For edges and contours, which are inherently linear or curvilinear features in 3D scenes, RANSAC enables the extraction of Gaussians that align well with the line segments while discarding outliers due to noise and artifacts from the optimization process. 

\textit{Robust selection of Sketch Gaussians relative to attributes}. Given a set of Sketch Gaussians associated with a 3D line, to account for the heterogeneous nature of Gaussian attributes, we perform RANSAC separately for each attribute (opacity, color, scaling, and rotation), thereby leveraging the unique characteristics of each.
The independent RANSAC results for opacity, color, scaling, and rotation are combined by intersecting the sets of inliers across all attributes. 
This ensures that only Gaussians consistent across multiple attributes are retained, improving the robustness and accuracy of the model to be estimated.

\textit{Robust selection of Sketch Gaussians relative to the model}. Specifically, for each attribute, we fit a polynomial regression model $f(\mathbf{\mu})$ to the robustly selected Sketch Gaussians along a given 3D line (see \cref{subsec:encode_optimize}). We then evaluate the residual error for each Gaussian as the squared deviation between its attribute value and the model’s prediction.  
Gaussians are classified as inliers if their residual error falls below a dynamically determined threshold 
\begin{equation}
    \epsilon = \eta \cdot \text{MAD},    
\end{equation}
where MAD denotes the Median Absolute Deviation, and $\eta$ is a hyperparameter that controls the sensitivity of RANSAC to deviations. Through this process, we identify a refined, robust and coherent set of Sketch Gaussians for each line segment. This set of Gaussian splats is represented efficiently through a compact polynomial model of the attributes. 

\textit{Generated Sketch Gaussians filtering}. Even though the two previous steps filter outliers within Sketch Gaussians along a 3D line, we observe that the scaling model tends to introduce a bias, which can result in poorly modeled long and thin Gaussians. 
We introduce a further post-processing phase to address a critical issue caused by the polynomial regression for scaling. 
To mitigate this, we decode the Sketch Gaussians and use the Interquartile Range (IQR) method to identify outlier Gaussians with extreme scaling values that deviate from the expected distribution. 
We then filter out the Gaussians among them that are not aligned with the direction of the line segment.
This post-processing ensures that only visually consistent Gaussians are retained for the later encoding, improving the accuracy of the Sketch Gaussian representation and ensuring better alignment with the underlying scene structure. 

Gaussians that are filtered out during this process are not discarded but instead reclassified as Patch Gaussians, ensuring that no geometric information is lost while maintaining the most appropriate representation for each Gaussian based on its characteristics.

\subsection{Gaussian Encoding and Optimization} \label{subsec:encode_optimize}

\textbf{Sketch Gaussian Encoding}.
After identifying the Sketch Gaussians, we aim to encode their attributes efficiently while preserving visual quality. Thanks to the coherence of the set of Sketch Gaussians along a 3D line, strengthened by the different robust steps detailed in the previous section, we compactly code the Sketch Gaussians. 
To achieve this, we choose polynomial regression (PR) to model the Gaussians’ attributes as it is a straightforward yet effective method for encoding complex features in a compact form. 
While more sophisticated methods could be used for encoding, PR combines simplicity with efficiency, which is crucial for real-time performance and memory consumption. 

For each line corresponding to a set of Sketch Gaussians, we train four PR models to encode their key attributes: opacity, color, scaling, and rotation. 
This separation of attributes allows for more accurate and flexible decoding and each model captures the relationships between the attribute and the line segments. 
The degree is chosen by grid search in the range of $[1,10]$. 
Once the models are trained, the attributes of the Sketch Gaussians are encoded as a parametric polynomial, significantly reducing the storage requirements. 

By encoding the Sketch Gaussians with PR models, we efficiently represent the boundary-defining features of the scene while maintaining visual fidelity. 
The compact representation not only reduces the memory footprint but also enables faster processing, making it a suitable approach for scalable 3D scene reconstruction and rendering. Our experiments (\cref{sec:experiments}), and in particular \cref{tab:gaussian_stat}, will stress out the efficiency of the Sketch Gaussians encoding.

\textbf{Patch Gaussian Pruning and Optimization}.
The core idea behind Patch Gaussian optimization is to reduce the number of Gaussians in smoother and broader regions of the 3D scene, where low-frequency variations are typically captured. 
While Sketch Gaussians efficiently encode high-frequency, boundary-defining features through 1D parametric models (where the original model concentrated significant storage), Patch Gaussians handle the volumetric representation of smoother regions where sparser distribution is sufficient. 
To this end, we leverage the concept of retraining, which allows for selective pruning of Patch Gaussians while ensuring that visual quality is maintained.

The process starts with the encoding and decoding of Sketch Gaussians. 
Once the Sketch Gaussians are (compactly) encoded, we decode them and fix the decoded version for subsequent operations, such as pruning and retraining. 
The set of decoded Sketch Gaussians remains unchanged during the pruning process. 
Note that the decoded Sketch Gaussians are used during retraining, rather than the original Sketch Gaussians, allowing the Patch Gaussians to compensate for errors introduced by the encoding-decoding process of the Sketch Gaussians.
The retraining focuses on optimizing the Patch Gaussians, which are first pruned, by randomly and uniformly removing some of them to improve the compactness of the model, and then retrained to improve the visual quality of the results.
This retraining process ensures that the Patch Gaussians are optimized to align with the visual structure of the scene \wrt the given set of Sketch Gaussians.

Through pruning and retraining, we achieve a more compact 3DGS representation where each Gaussian type serves its designated role: Sketch Gaussians efficiently capture linear features and boundaries, while optimized Patch Gaussians provide comprehensive coverage of smoother regions.
This dual optimization strategy significantly reduces the total number of Gaussians required to represent the scene while maintaining high fidelity. 
The result is a hybrid representation that efficiently allocates storage resources according to the geometric characteristics of different scene regions, \ie compact parametric models for boundary features and optimized sparse distribution for volumetric regions.



\add{
\textbf{Vector Quantization}.
For additional storage efficiency, we apply vector quantization to the Patch Gaussians following Papantonakis \etal~\cite{papantonakis2024reducing}. This compression scheme employs K-means clustering to create codebooks for various Gaussian attributes, including opacity, scaling, quaternion rotation (real and imaginary parts), base color coefficients, and spherical harmonics color components. Instead of storing exact values, the indices are maintained to the nearest values in fixed-size codebooks. For vector attributes, separate indices are used for each component while sharing a single codebook. Based on the empirical experiments from Papantonakis \etal~\cite{papantonakis2024reducing}, 256-entry codebooks with 1-byte indices are used, offering optimal compression while preserving visual quality. 
Note that Gaussian positions are not compressed with codebooks because it leads to significant quality degradation, according to their pilot experiments. Instead, 16-bit half-float quantization is applied to Gaussian positions and codebook entries, to further reduce storage requirements while maintaining visual fidelity.
}

\begin{figure*}[t!]%
\centering
    \subfloat[][Playroom.]{%
        \includegraphics[width=0.24\textwidth]{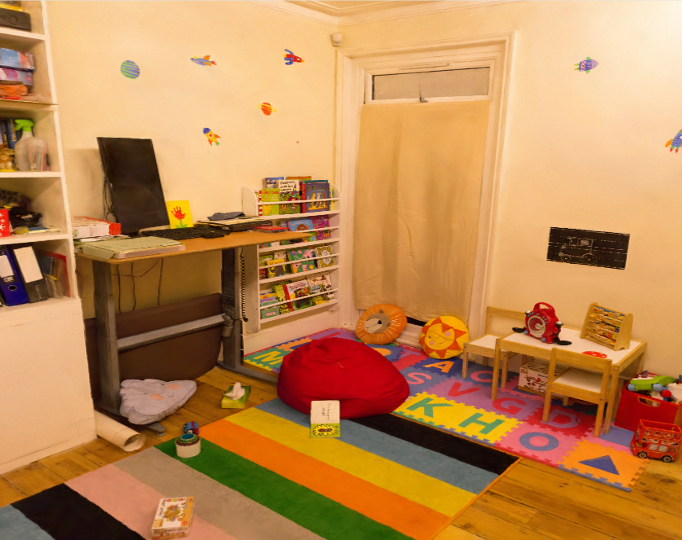}
        }
    \subfloat[][Drjohnson.]{%
        \includegraphics[width=0.24\textwidth]{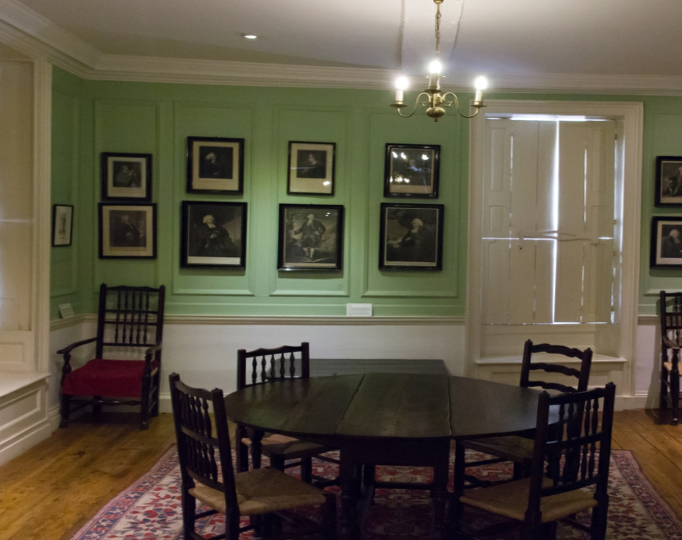}
        }
    \subfloat[][Room.]{%
        \includegraphics[width=0.24\textwidth]{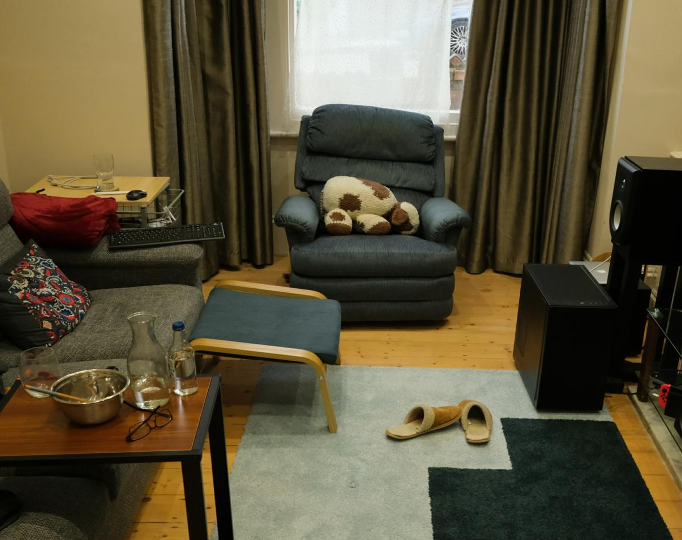}
        }
    \subfloat[][Truck.]{%
        \includegraphics[width=0.24\textwidth]{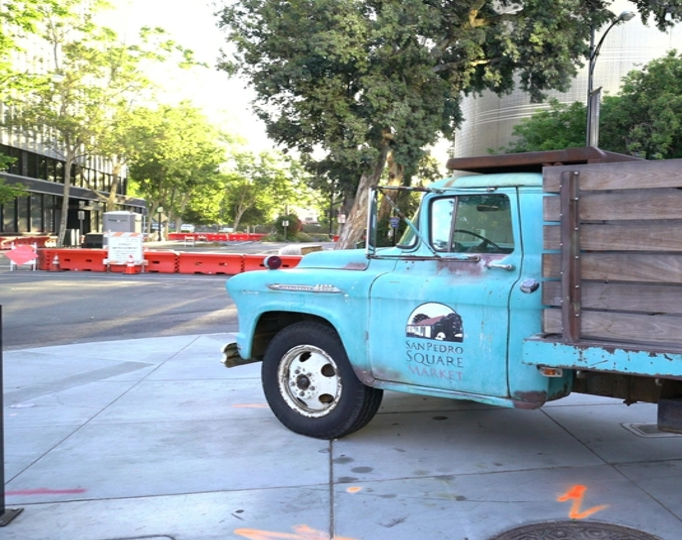}
        }
    \caption[]{The four scenes used in our evaluation: (a) Playroom, (b) Drjohnson, (c) Room, and (d) Truck. These scenes encompass diverse man-made environments, ranging from interior spaces with furniture and architectural details to exterior objects.}%
\label{fig:thumbnails}
\end{figure*}

\section{Experimental results}
\label{sec:experiments}

\subsection{Experimental Setup}

\textbf{Dataset and Metrics}.
We evaluated our method across four representative scenes from three distinct datasets: \textit{Playroom} and \textit{Drjohnson} from the Deep Blending dataset~\cite{hedman2018deep}, \textit{Room} from the Mip-NeRF360 dataset~\cite{barron2022mip}, and \textit{Truck} from the Tanks\&Temples dataset~\cite{knapitsch2017tanks}. 
As shown in \cref{fig:thumbnails}, these scenes were specially selected to encompass a diverse range of geometric complexities, including bounded indoor environments and expansive unbounded outdoor scenes. 
This selection allows for an assessment of our method's adaptability across varying scene morphologies characteristic of man-made environments.

We quantitatively assessed the visual fidelity of our 3D Gaussian representations using three complementary metrics: peak signal-to-noise ratio (PSNR), structural similarity index measure (SSIM)~\cite{wang2004image}, and learned perceptual image patch similarity (LPIPS)~\cite{zhang2018unreasonable}.

\textbf{Implementation}.
Our implementation is based on the official 3D Gaussian Splatting codebase~\cite{gaussiancode}. 
We initially trained 3DGS models for each scene using the default hyperparameters on an NVIDIA H100 GPU. 
Subsequently, we employed Line3D++ using the official release code~\cite{line3dcode} to extract 3D line segments, which formed the foundation for identifying Sketch Gaussians.
After distinguishing the Sketch and Patch Gaussians with the line prior, the polynomial regression (PR) is utilized to model Sketch Gaussians while Patch Gaussians are pruned, retrained, and quantized for further compact representation. 
The PR model's degree was determined through a comprehensive grid search, ensuring optimal representation. 
We maintained consistent hyperparameters across all training stages. 

\add{The identification and modeling of Sketch Gaussians were implemented using Python's multiprocessing capabilities, running on dual AMD Epyc 9334 CPUs with 3.9 GHz and 32 cores per CPU. Our experiments show that through parallel processing optimization, we can achieve $7\times$ speedup on average, taking only \qtyrange[range-units=single]{5}{10}{\minute} when processing $1,000-2,000$ line segments per scene.
The subsequent Patch Gaussian retraining takes a fraction of the original 3DGS training time. Therefore, our method could achieve comparable or even faster processing times compared to standard 3DGS training while producing more storage-efficient representations.}

\textbf{Comparison Methods}.
We implemented three comparative approaches to validate our method:
\begin{itemize}
    \item \textit{Baseline}. We train 3DGS for each scene with different densification thresholds, producing multiple 3DGS models with different sizes. 
    Standard 3DGS relies primarily on 2D positional gradients to guide the densification of Gaussians, which frequently results in significant redundancy, particularly in boundary-defining regions. 
    Serving as the baseline, this approach globally adapts the densification gradient by changing the threshold value. We show that a global threshold modification fails to address the fundamental \gm{redundancy} in Gaussian representation. 
    \item \textit{Sketch}. The method focuses exclusively on the encoding of Sketch Gaussians, deliberately omitting the pruning and retraining of Patch Gaussians. 
    To quantify the storage efficiency and visual quality trade-offs inherent in compact boundary feature representation, we progressively reduce the number of line segments used for Sketch Gaussian identification and modeling and measure the visual quality and model size. Note that this method is the only one not to retrain the model.
    \item \textit{Prune\&Retrain}. 
    This method applies a uniform pruning and retraining strategy across all Gaussians, without distinguishing between Sketch and Patch categories. 
    This method serves as an ablation study to demonstrate the importance of recognizing and processing Gaussians differently based on their structural roles.
\end{itemize}

\begin{figure*}[t]%
\centering
    \subfloat[][Playroom.]{%
        \includegraphics[width=0.24\textwidth]{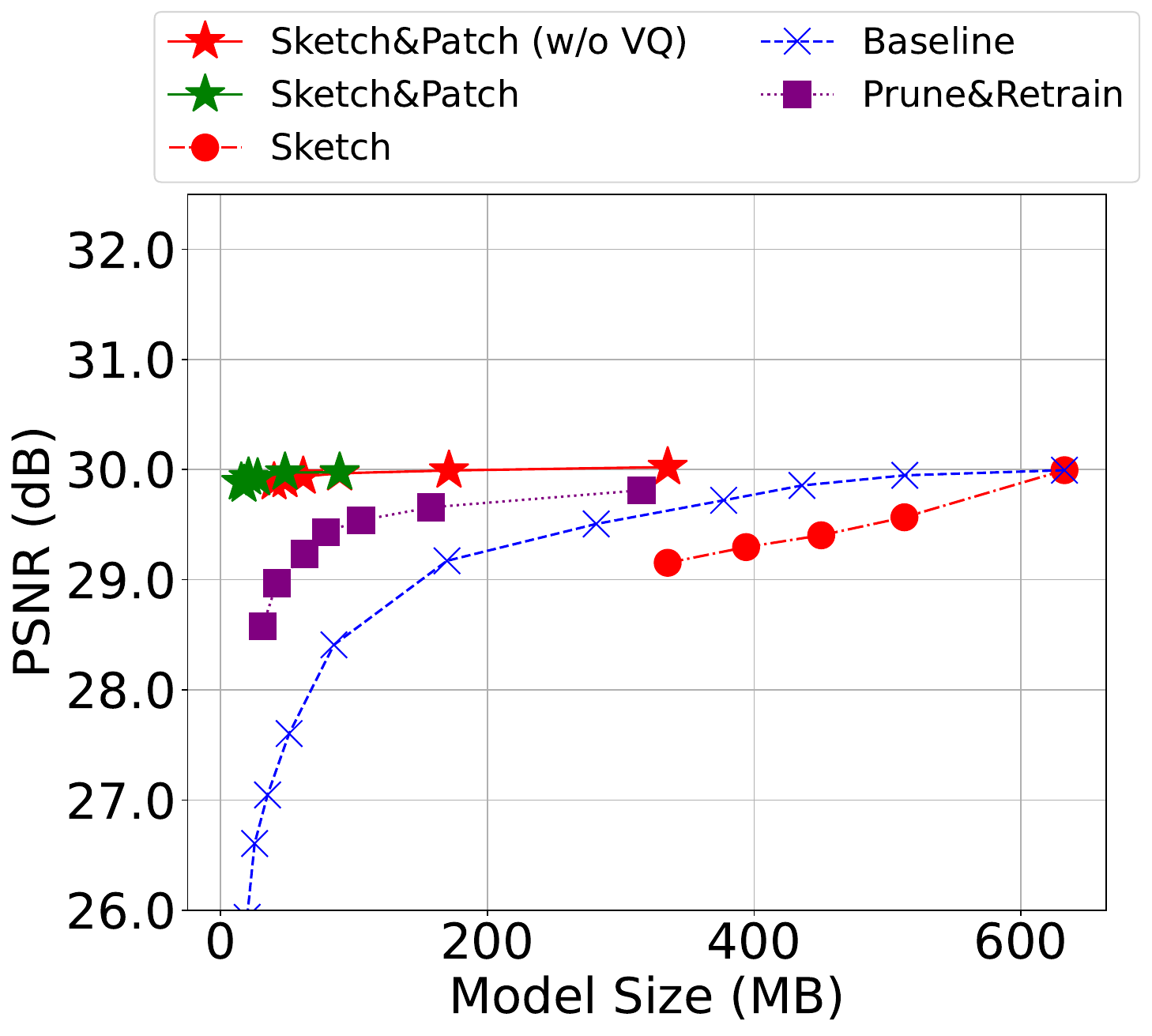}
        }
    \subfloat[][Drjohnson.]{%
        \includegraphics[width=0.24\textwidth]{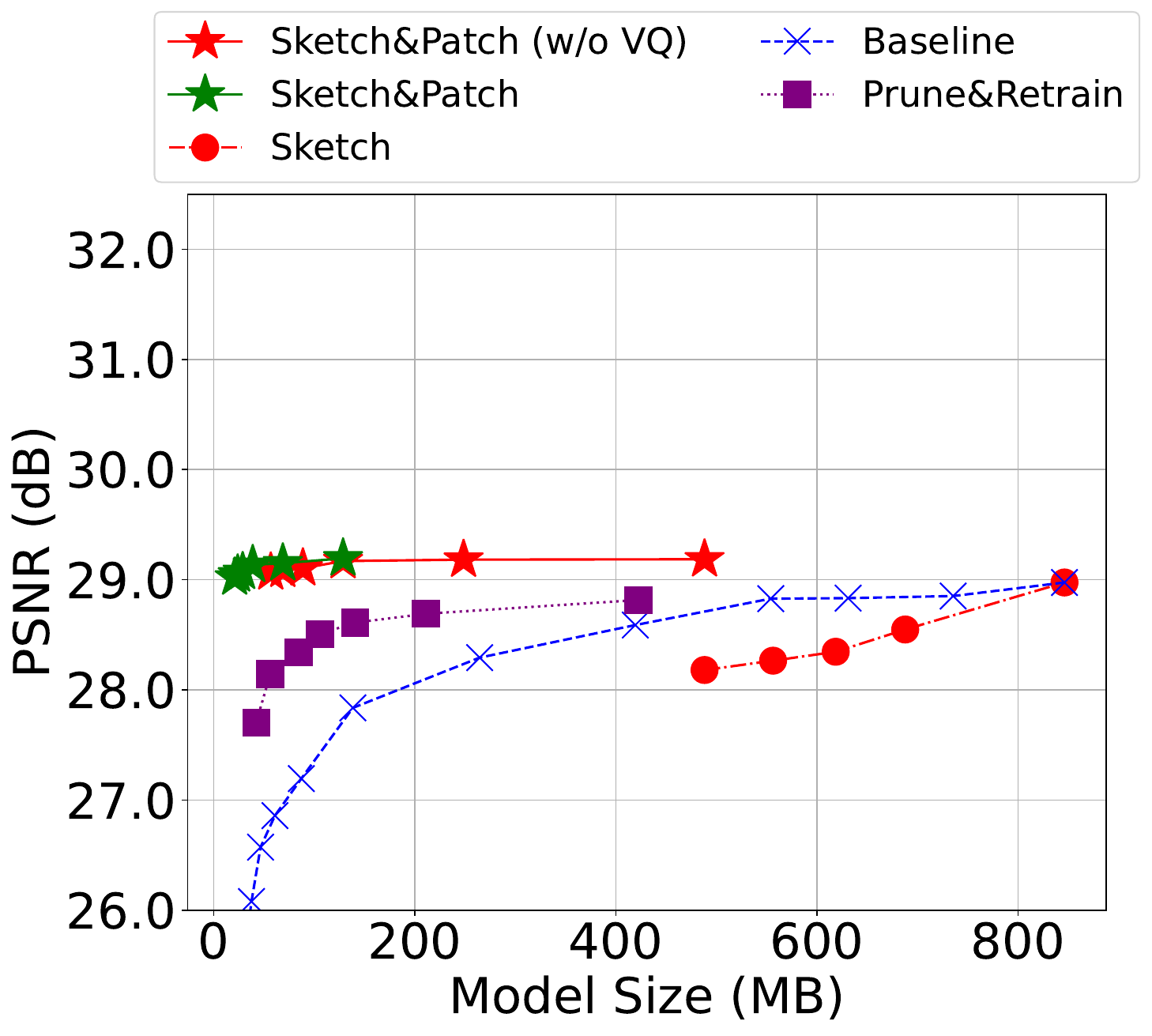}
        }
    \subfloat[][Room.]{%
        \includegraphics[width=0.24\textwidth]{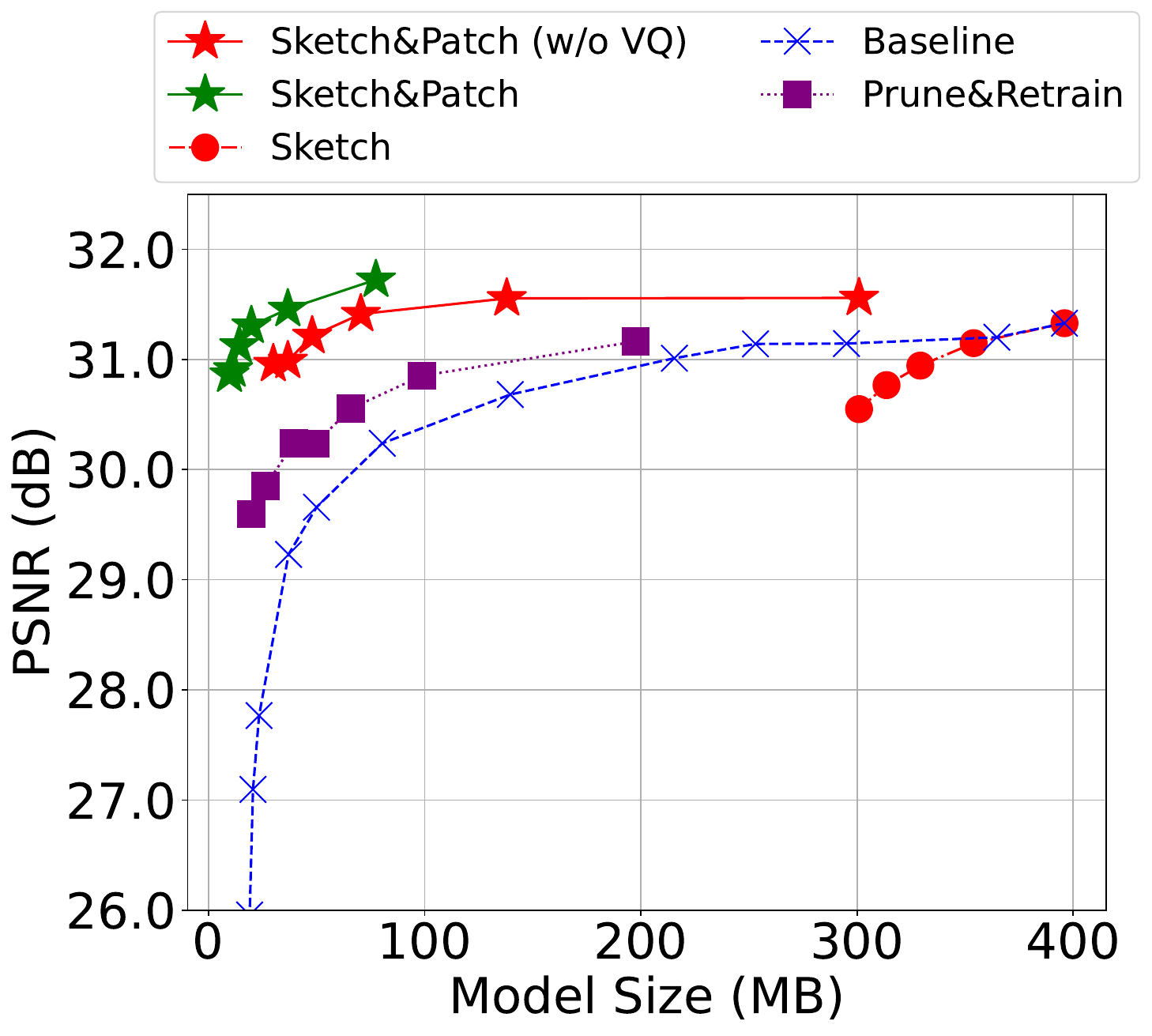}
        }
    \subfloat[][Truck.]{%
        \includegraphics[width=0.24\textwidth]{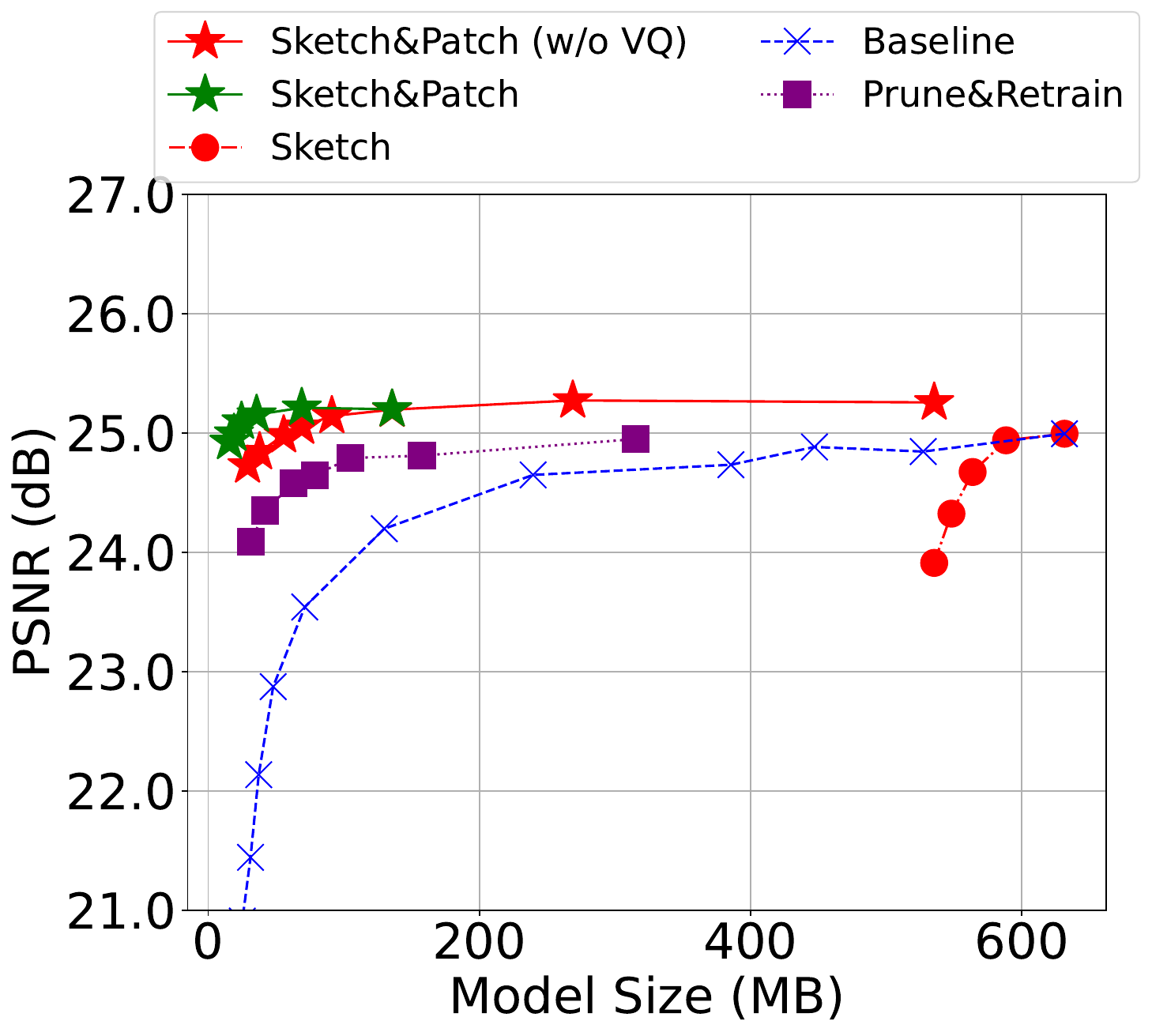}
        }
    \caption[]{R-D curves for PSNR: (a) Playroom, (b) Drjohnson, (c) Room, and (d) Truck.}%
    \label{fig:psnr_rd_curve}
    \vspace{0.6cm}
    \centering
    \subfloat[][Playroom.]{%
        \includegraphics[width=0.24\textwidth]{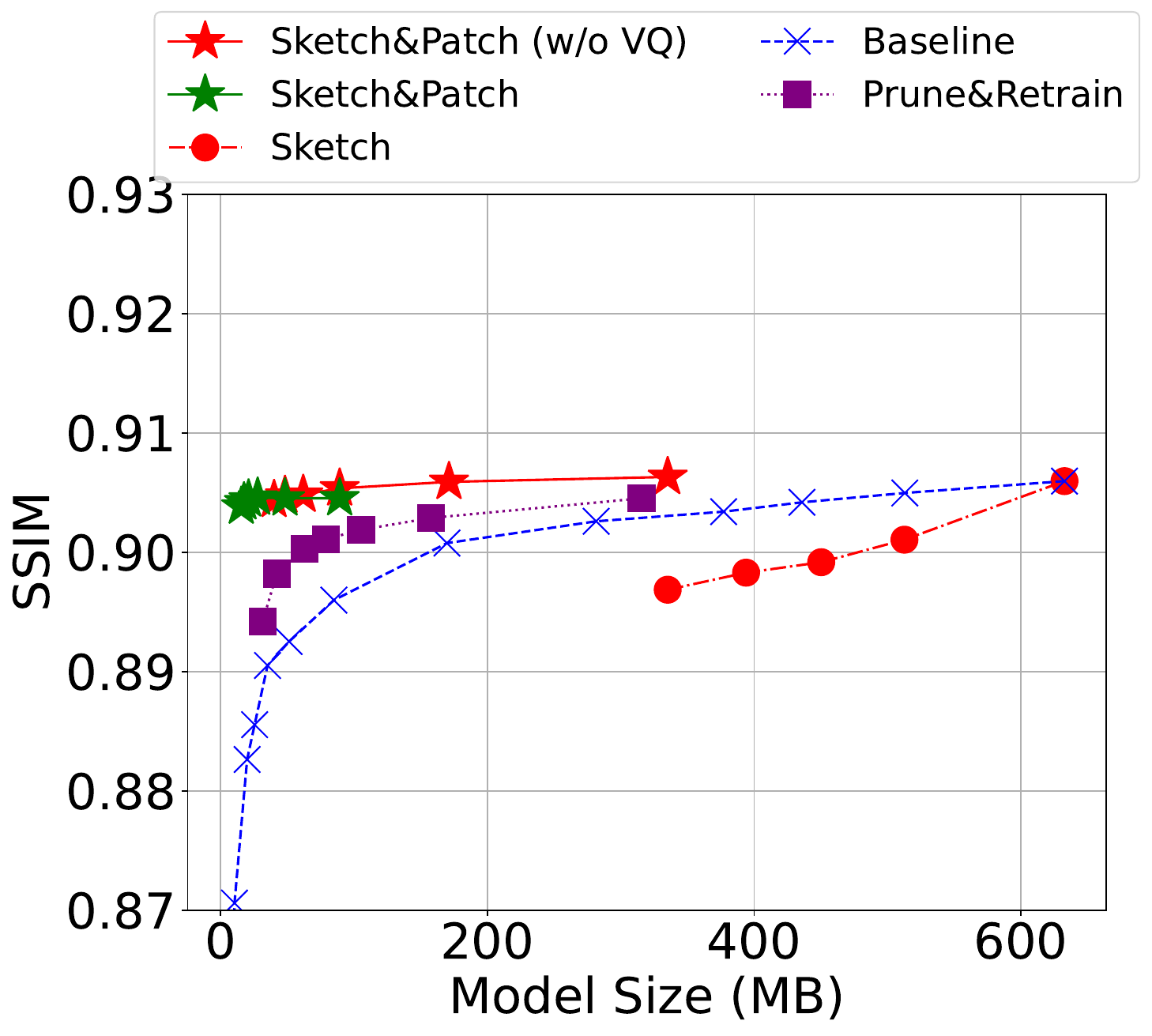}
        }
    \subfloat[][Drjohnson.]{%
        \includegraphics[width=0.24\textwidth]{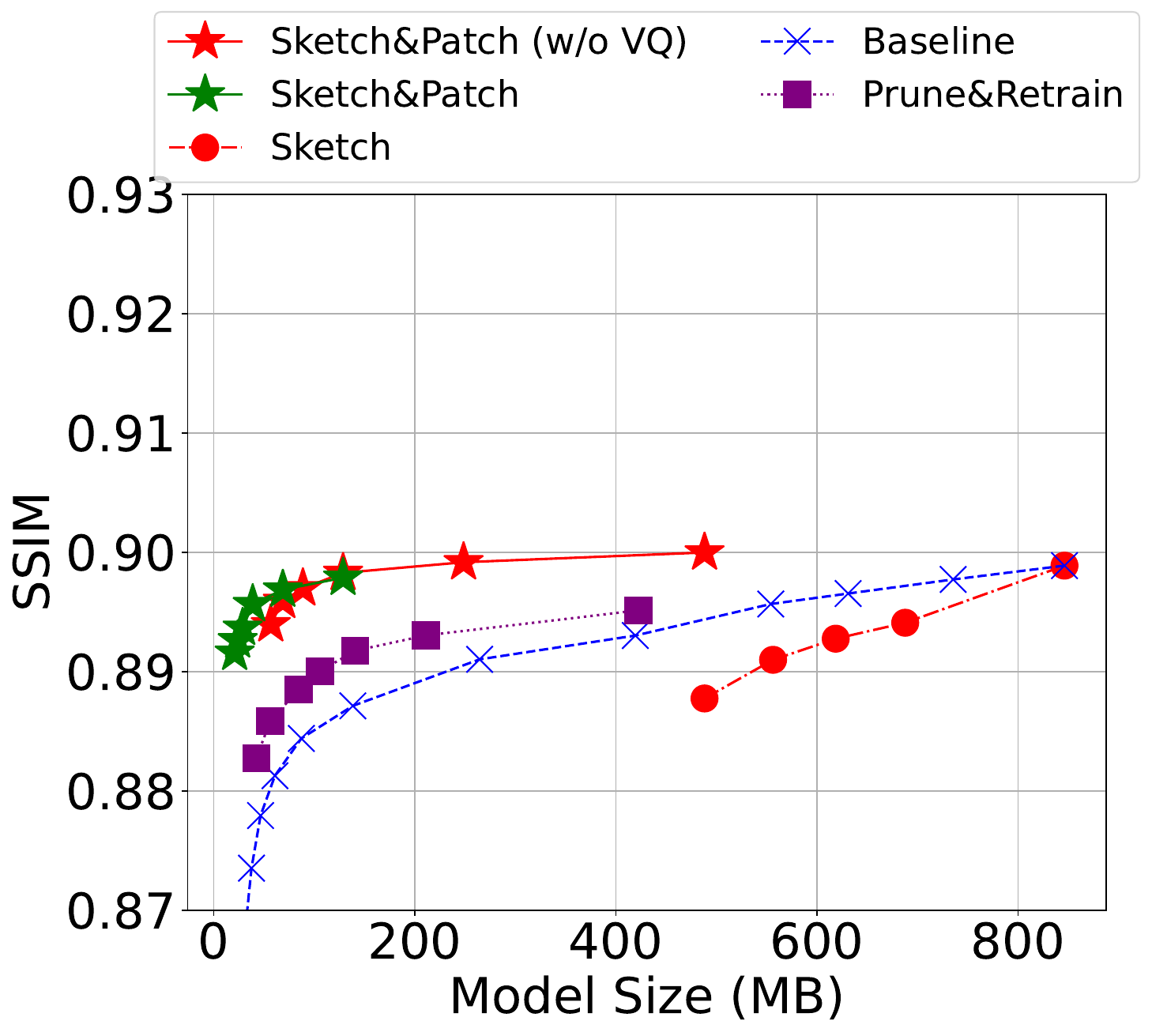}
        }
    \subfloat[][Room.]{%
        \includegraphics[width=0.24\textwidth]{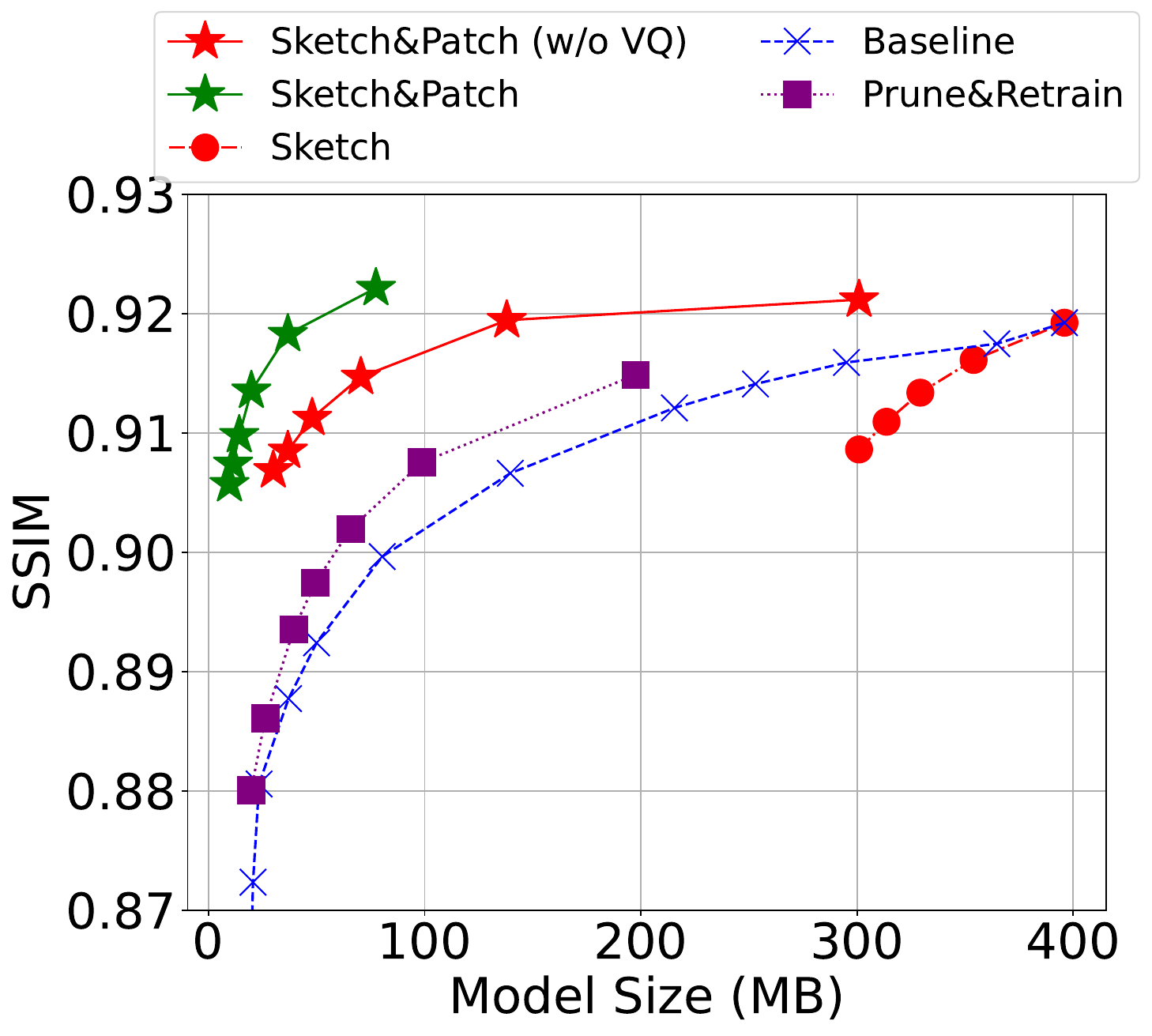}
        }
    \subfloat[][Truck.]{%
        \includegraphics[width=0.24\textwidth]{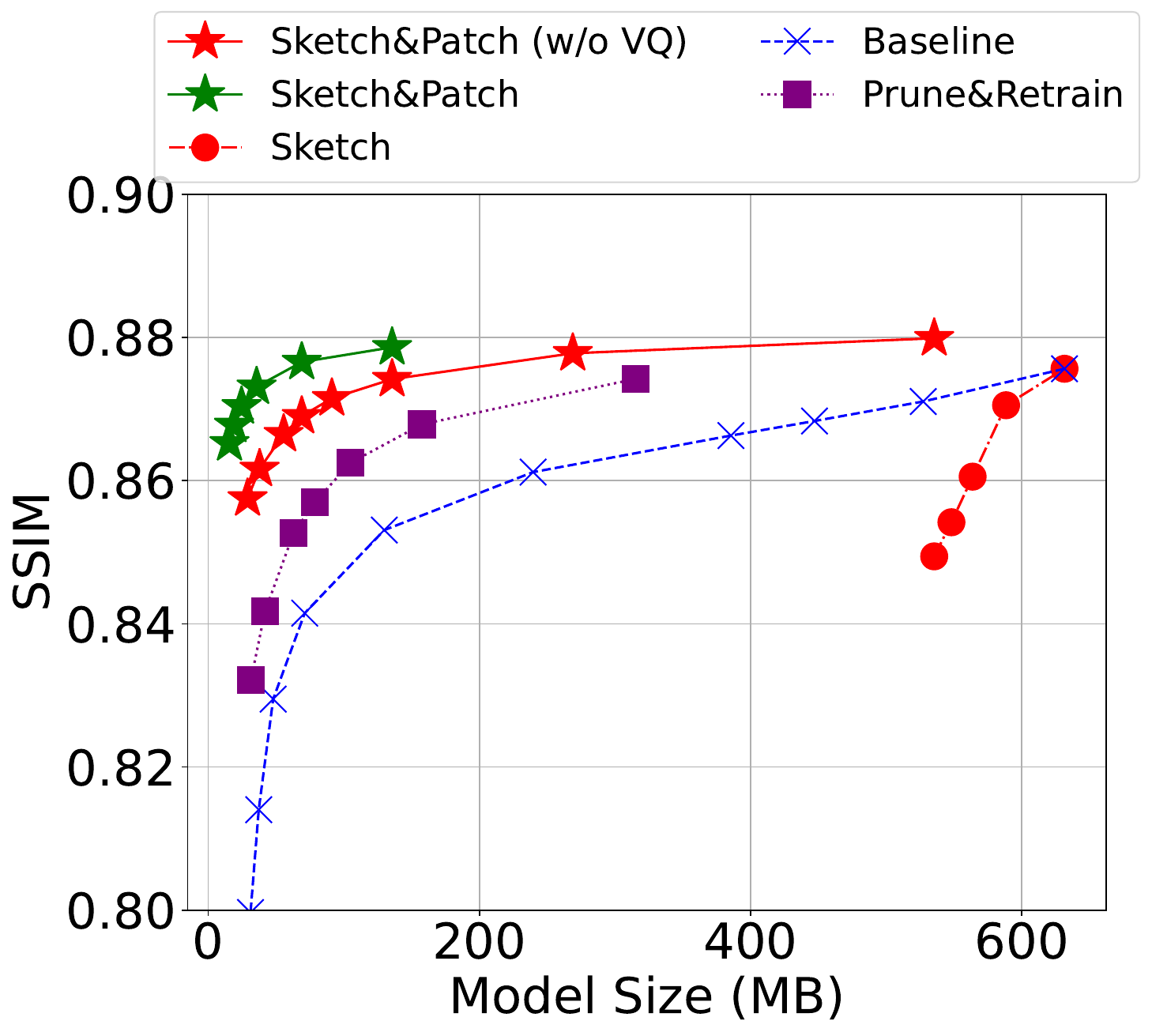}
        }
    \caption[]{R-D curves for SSIM: (a) Playroom, (b) Drjohnson, (c) Room, and (d) Truck.}%
    \label{fig:ssim_rd_curve}
    \vspace{0.6cm}
    \centering
    \subfloat[][Playroom.]{%
        \includegraphics[width=0.24\textwidth]{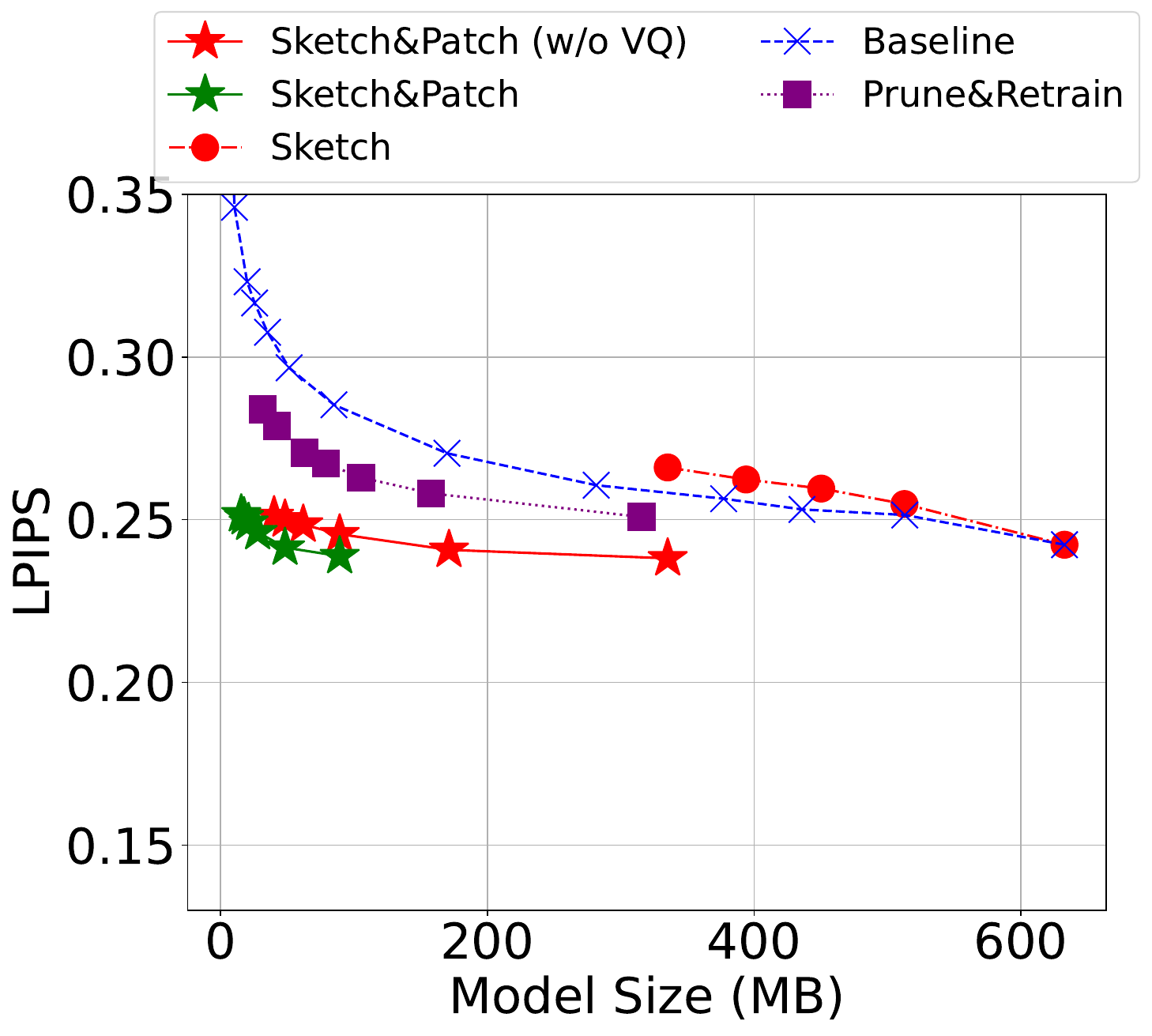}
        }
    \subfloat[][Drjohnson.]{%
        \includegraphics[width=0.24\textwidth]{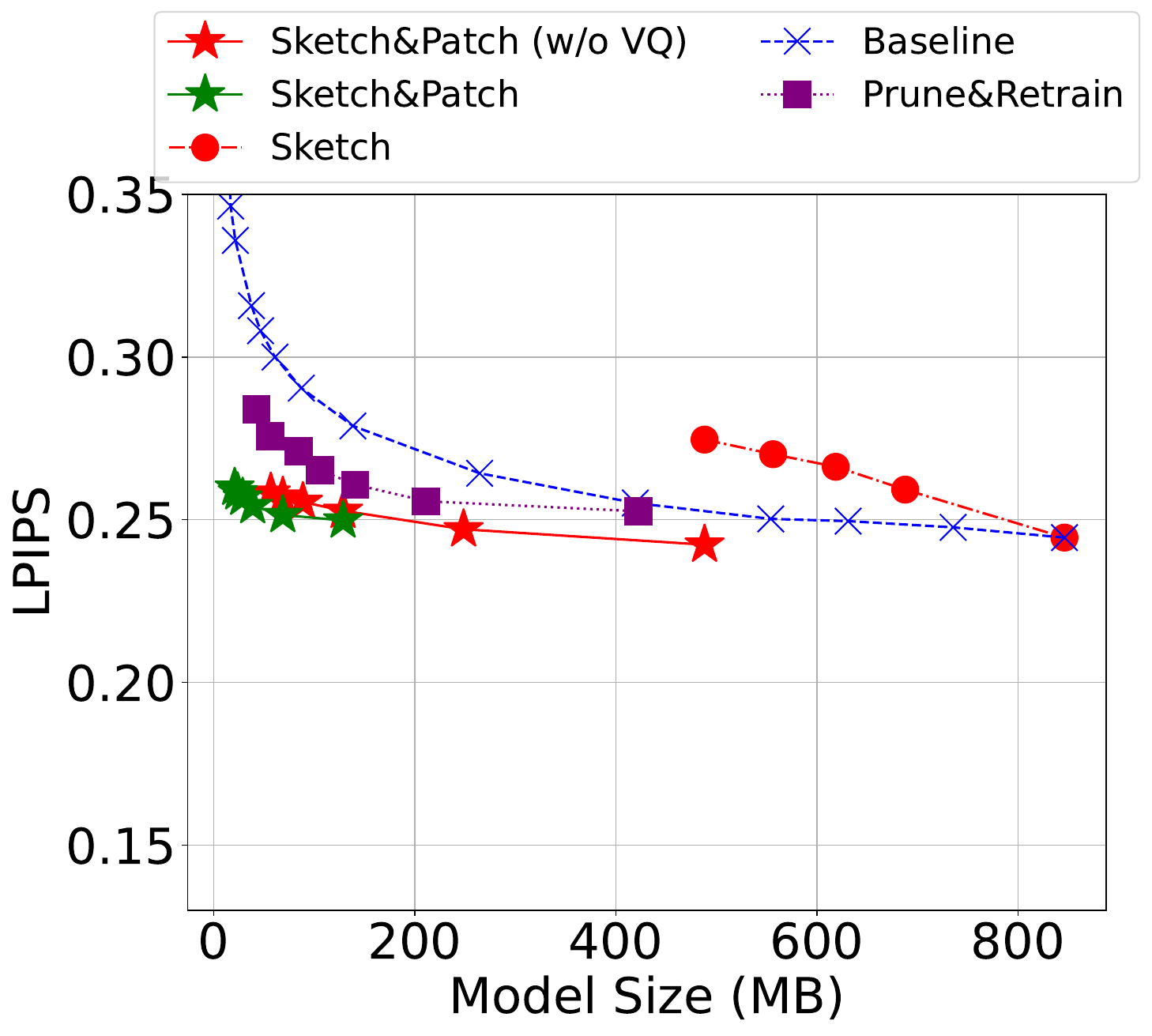}
        }
    \subfloat[][Room.]{%
        \includegraphics[width=0.24\textwidth]{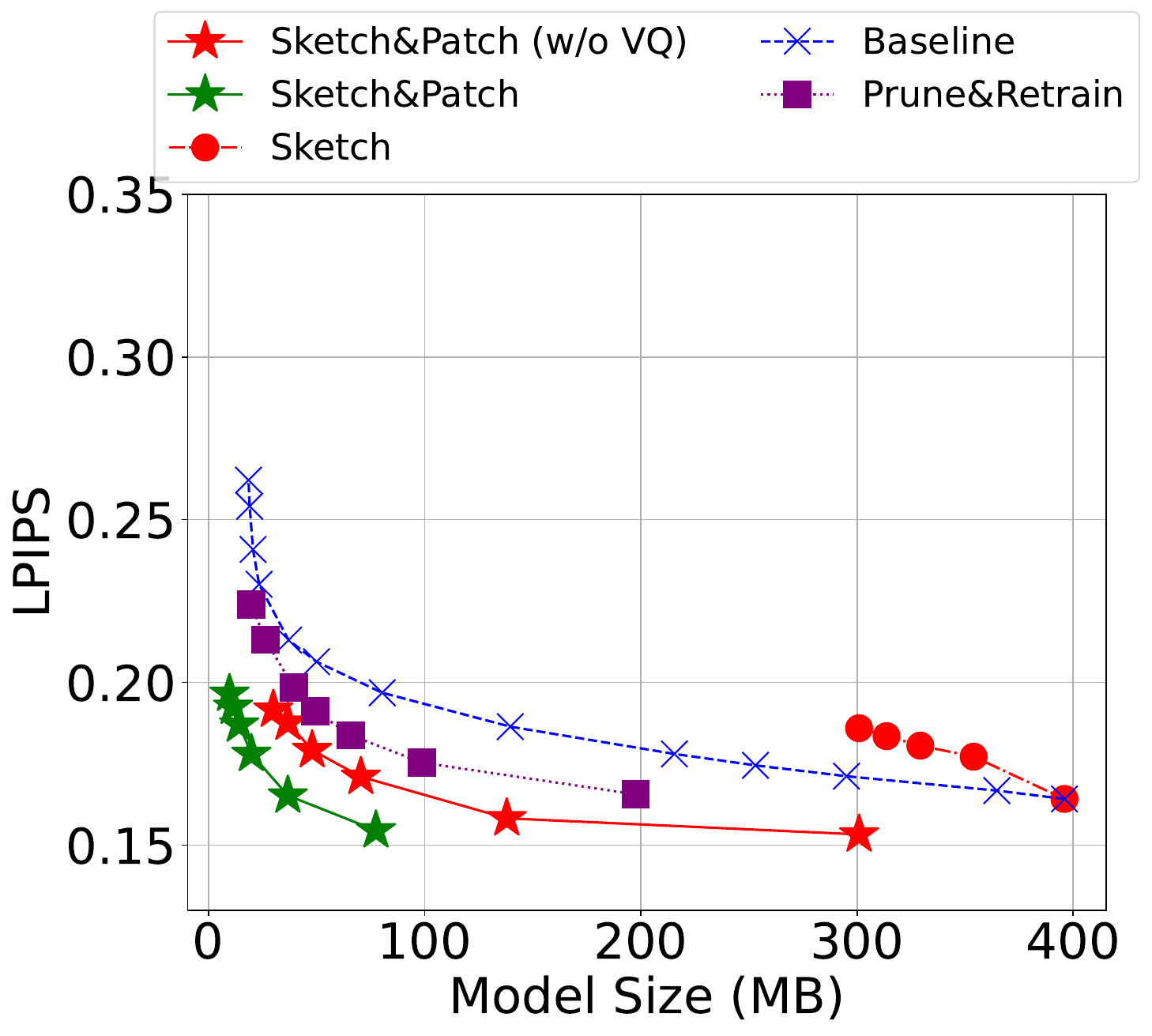}
        }
    \subfloat[][Truck.]{%
        \includegraphics[width=0.24\textwidth]{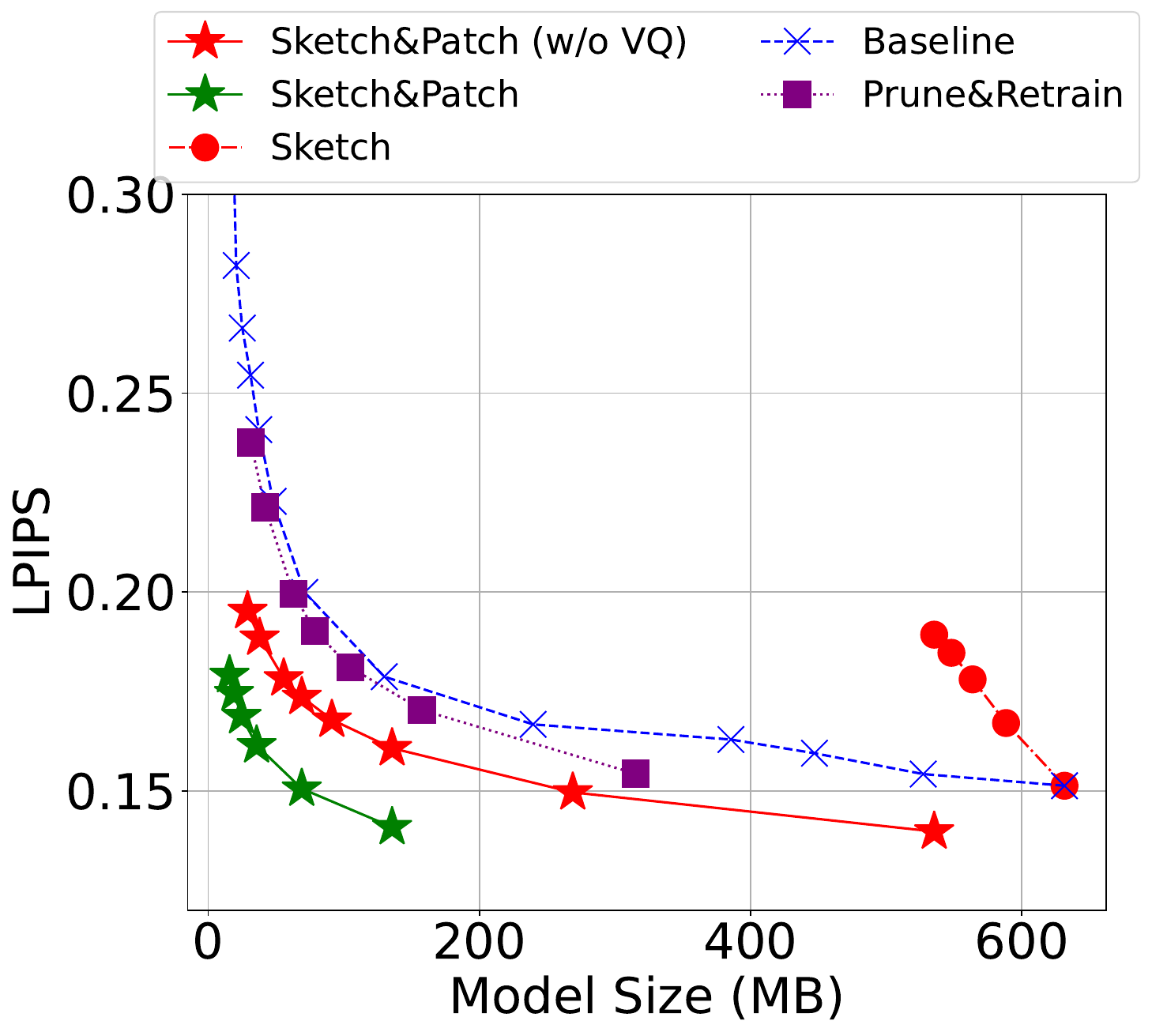}
        }
    \caption[]{R-D curves for LPIPS: (a) Playroom, (b) Drjohnson, (c) Room, and (d) Truck.}%
    \label{fig:lpips_rd_curve}
\end{figure*}

\subsection{Results Analysis}

\cref{fig:psnr_rd_curve,fig:ssim_rd_curve,fig:lpips_rd_curve} plot the visual quality of the four scenes versus the model size for four methods.
To generate rate-distortion (R-D) curves, we varied parameters for each method. 
For the \textit{Baseline}, we adjusted the densification gradient threshold from $0.002$ (default) to $0.2$ with an interval of $0.0005$. For the \textit{Prune\&Retrain}, we created multiple models by reducing the number of Gaussian splats by factors of $\{2, 4, 6, 8, 10, 15, 20\}$ through uniform pruning, followed by retraining. 
For the \textit{Sketch} method, we varied line segments used for Sketch Gaussian identification by selecting the top $\{\SI{25}{\percent}, \SI{50}{\percent}, \SI{75}{\percent}, \SI{100}{\percent}\}$ longest lines. 
As for our proposed method (denoted as \textit{Sketch\&Patch}), we first utilized all line segments for Sketch Gaussian identification and modeling, then reduced Patch Gaussians by factors of $\{2, 4, 6, 8, 10, 15, 20\}$ through uniform pruning and retraining.

We make the following observations from these results.

\textbf{Performance in Visual Quality}. 
First, our proposed \textit{Sketch\&Patch} method demonstrates superior performance across all evaluated metrics compared to alternative approaches at the same model size. 
For instance, in the Playroom scene, we observe enhancements by up to \SI{18.96}{\percent} in PSNR, \SI{3.12}{\percent} in SSIM, and \SI{24.72}{\percent} in LPIPS.
Similar patterns of improvement are evident in other scenes, with Drjohnson showing gains of \SI{16.27}{\percent} in PSNR, \SI{3.65}{\percent} in SSIM, and \SI{22.62}{\percent} in LPIPS at most.
Our method on Room scene achieves improvements by up to \SI{22.78}{\percent} in PSNR, \SI{6.80}{\percent} in SSIM, and \SI{25.03}{\percent} in LPIPS, while the Truck scene demonstrates enhancements of \SI{32.62}{\percent} in PSNR, \SI{19.12}{\percent} in SSIM, and \SI{45.41}{\percent} in LPIPS.
These substantial improvements can be attributed to our method's strategic approach to Gaussian representation. 
By differentiating between Sketch and Patch Gaussians, our method preserves critical geometric features while efficiently representing broader scene regions. 
The particularly high performance on perceptual metrics (LPIPS) suggests that our approach effectively maintains the visually significant features contributing to human perception of scene quality.

\textbf{Storage Efficiency}.
Second, a key advantage of our method lies in its significant reduction in storage requirements while maintaining visual fidelity. 
When comparing model sizes at equivalent SSIM values, for instance, our method achieves remarkable efficiency: requiring only \SI{2.35}{\percent} of the original model size for Playroom, \SI{5.41}{\percent} for Drjohnson, \SI{4.83}{\percent} for Room, and \SI{3.46}{\percent} for Truck, while maintaining the same visual quality.
These substantial reductions in storage requirements demonstrate the effectiveness of our hybrid representation strategy.
The achieved compression rates can be attributed to two key factors: the efficient encoding of high-frequency features through parametric models (Sketch Gaussians encoding) and the optimized representation of smooth regions through carefully pruned and quantized Patch Gaussians.
This dual approach enables our method to maintain high visual quality while significantly reducing the storage footprint, addressing a critical challenge in 3D scene representation.

\textbf{Ablation Effect of Sketch and Patch}. 
As shown in \cref{fig:psnr_rd_curve,fig:ssim_rd_curve,fig:lpips_rd_curve}, while the \textit{Sketch} method
does not surpass the baseline's quality-storage trade-off, it achieves comparable performance. 
This observation is particularly noteworthy given that the \textit{Sketch} method operates without the benefit of Patch Gaussian optimization. 
The competitive performance of this naive approach validates our fundamental hypothesis regarding the high coherence of boundary-defining Gaussians and their amenability to efficient encoding.

Moreover, the superior performance of our complete method over the \textit{Prune\&Retrain} approach underscores the importance of differential treatment of Gaussian categories. 
As shown, with the same model size, our method consistently achieves better visual quality. 
This is because uniform pruning and retraining, while effective for general optimization, fail to capitalize on the distinct characteristics of boundary and non-boundary regions.

\begin{figure}[t]
    \centering
    \includegraphics[width=0.45\textwidth]{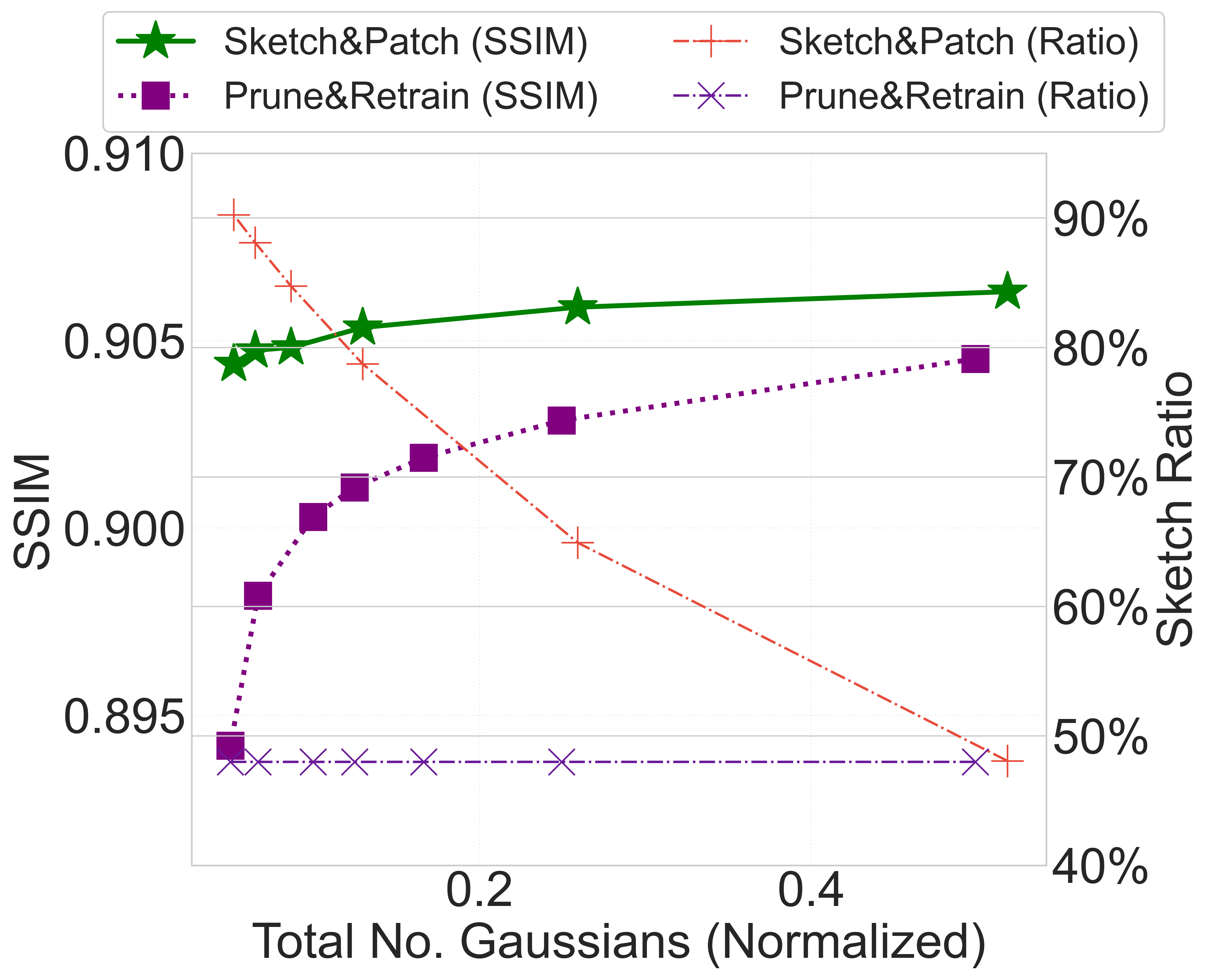}
    \caption{Visual quality (SSIM) vs. ratio of Sketch Gaussians of scene \textit{Playroom} constructed by \textit{Sketch\&Patch}.}
    \label{fig:sketch_ratio_plot}
\end{figure}

To better illustrate it, we show the visualization of the rendering performance vs. the ratio of Sketch Gaussians of the scene \textit{Playroom} constructed by method \textit{Sketch\&Patch} in \cref{fig:sketch_ratio_plot}. 
The Sketch ratio, defined as the number of Sketch Gaussians over the total number of Gaussian splats, provides valuable insight into the effectiveness of our method. 
As can be observed, in our \textit{Sketch\&Patch} method, the Sketch ratio increases as the model size decreases, ranging from approximately \SI{48}{\percent} at the largest model size to \SI{90}{\percent} at the most compressed state. 
This trend occurs because we fix the Sketch Gaussians while progressively pruning and optimizing the Patch Gaussians, ensuring the preservation of critical boundary features even under extreme compression.
In contrast, the \textit{Prune\&Retrain} method maintains a constant Sketch ratio of approximately \SI{48}{\percent} across all model sizes, as it uniformly prunes and optimizes all Gaussian splats without distinguishing their roles. 
This uniform treatment leads to suboptimal resource allocation, particularly evident in their SSIM values. 
Even when our method has a high Sketch ratio ($> \SI{80}{\percent}$) at small model sizes, it achieves better visual quality compared to \textit{Prune\&Retrain}. 
This superior performance demonstrates that preserving Sketch Gaussians while selectively optimizing Patch Gaussians is more effective than uniform pruning, especially under stringent storage constraints.
In summary, the consistent superiority of our method validates the significance of distinguishing between Sketch and Patch Gaussians, enabling targeted optimization strategies for each category.


\begin{table}[h]
        \caption{Model size (MB) at equivalent visual quality level.} 
        \label{tab:gaussian_stat}
        \centering
            \begin{tabular}{l|cccc}
                \toprule
                \textbf{Model} & \textbf{Playroom} & \textbf{Drjohnson} & \textbf{Room} & \textbf{Truck}\\
                \midrule
                Vanilla 3DGS  & $631.44$ & $844.48$ & $395.16$ & $630.23$\\
                Sketch-Patch GS & $19.07$ & $41.46$ & $21.05$ & $36.59$ \\
                \midrule
                \hspace{0mm}Patch Gaussians & $328.35$ & $481.35$ & $268.81$ & $535.70$ \\
                \hspace{2mm}+ Pruned and Retrained & $46.90$ & $127.36$ & $69.93$ & $135.28$ \\
                \hspace{2mm}+ Vector Quantization & $11.72$ & $31.84$ & $17.48$ & $33.82$ \\
                \hspace{0mm}Sketch Gaussians & $303.09$ & $363.13$ & $126.45$ & $94.53$\\
                \hspace{2mm}+ Encoded with PR & $7.35$ & $9.62$ & $3.57$ & $2.77$\\
                \bottomrule
            \end{tabular}
\end{table}

\textbf{Breakdown of Components of Our Method}.
We further present \cref{tab:gaussian_stat}, which provides a detailed breakdown of our hybrid representation's storage requirements, comparing Sketch and Patch Gaussians components with the baseline Vanilla 3DGS at an equivalent visual quality level in terms of SSIM ($0.88$ on Truck, $0.92$ on Room, $0.90$ on Drjohnson, and $0.91$ on Playroom).

The results reveal that our parametric encoding of Sketch Gaussians achieves remarkable efficiency, requiring no more than \SI{9.62}{\mega\byte} of storage across all test scenes. 
This significant reduction in storage requirements for Sketch Gaussians demonstrates the effectiveness of leveraging geometric coherence through parametric models.
The compact representation of Sketch Gaussians is particularly noteworthy, as these elements typically require dense sampling in traditional approaches to maintain edge and contour fidelity. 
Our method's ability to encode these features efficiently while preserving their visual importance validates the effectiveness of our line segment-based parametric approach. 
Furthermore, this efficiency in Sketch Gaussian representation allows for a more optimal allocation of storage resources to Patch Gaussians, contributing to the overall balance between storage efficiency and visual quality.

Moreover, we can observe that pruning\&optimization and vector quantization contribute significantly to compressing the Patch Gaussians without impacting the visual quality. Specifically, applying pruning and optimization reduces the size of the Patch Gaussians $5\times$ while vector quantization further cuts down the storage of pruned Patch Gaussians $4\times$, resulting in an average of $19\times$ reduction in size.

\begin{figure*}[t]%
    \centering
    \includegraphics[width=\textwidth]{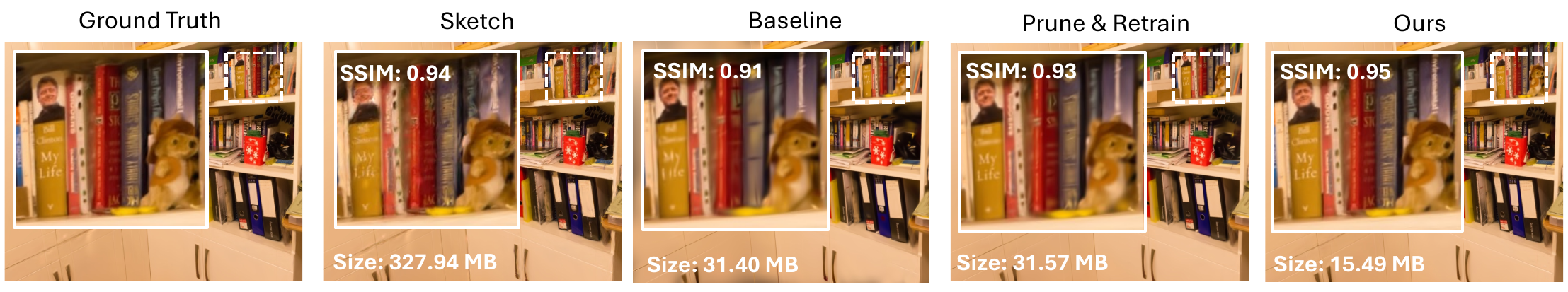}
    \caption[]{Rendered images of 3DGS on \textit{Playroom} generated from four methods. }%
    \label{fig:vis_playroom}
    \vspace{0.6cm}
    \centering
    \includegraphics[width=\textwidth]{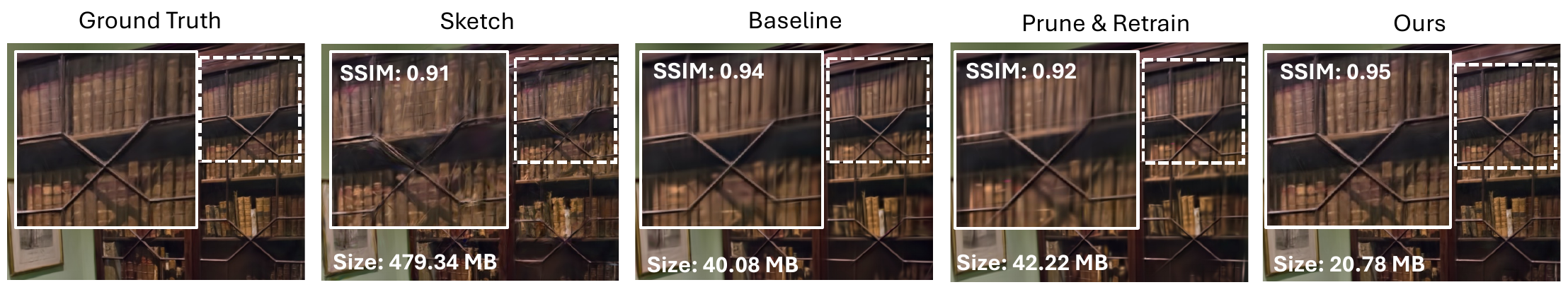}
    \caption[]{Rendered images of 3DGS on \textit{Drjohnson} generated from four methods.}%
    \label{fig:vis_drjohnson}
    \vspace{0.6cm}
    \centering
    \includegraphics[width=\textwidth]{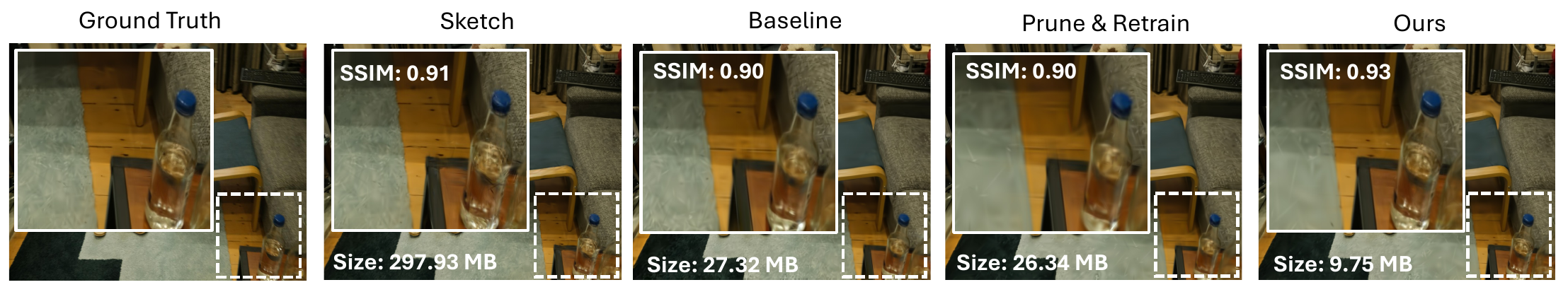}
    \caption[]{Rendered images of 3DGS on \textit{Room} generated from four methods.}%
    \label{fig:vis_room}
    \vspace{0.6cm}
    \centering
    \includegraphics[width=\textwidth]{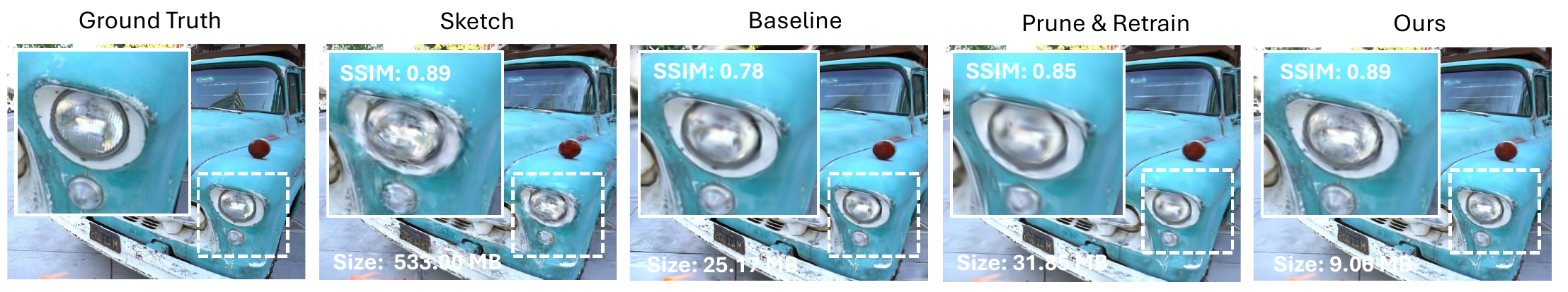}
    \caption[]{Rendered images of 3DGS on \textit{Truck} generated from four methods.}%
    \label{fig:vis_truck}
\end{figure*}

\textbf{Qualitative Comparison}.
To illustrate the performance of the proposed method, we also present the visual comparison with the other methods in \cref{fig:vis_playroom,fig:vis_drjohnson,fig:vis_room,fig:vis_truck}, for qualitative analysis. 
For fair comparison, we selected models representing each method's maximum achievable compression while maintaining acceptable visual quality, corresponding to the leftmost points in the R-D curves. 

It can be noted that our method excels particularly in preserving geometric details and structural integrity. 
Compared to \textit{Baseline} and \textit{Prune\&Retrain} method, the Sketch Gaussian encoding effectively preserves sharp edges and boundary features, especially in challenging areas such as bookshelves with stacked books and clearly readable book titles. 
This preservation of fine text and sharp edges is particularly impressive given the substantial reduction in storage requirements. 
Simultaneously, the optimized Patch Gaussians successfully preserve surface continuity in broader regions, eliminating redundant information while maintaining visual consistency.

In contrast, the \textit{Prune\&Retrain} method, while achieving moderate compression, exhibits noticeable degradation in geometric detail preservation, particularly evident in the degradation of text legibility and book edge definition. The Sketch-only approach, despite its effectiveness in boundary preservation through parametric modeling, fails to achieve significant storage savings without the pruning of Patch Gaussians.

\section{Discussion and Conclusion}
\label{sec:conclude}

This paper introduces a novel perspective on 3D Gaussian scene representation by recognizing and leveraging the distinct roles of different Gaussian types in man-made environments. 
Experimental results show that the proposed hybrid representation significantly reduces storage requirements while maintaining visual quality, addressing a critical challenge in immersive media applications.

While our method shows promising results, there are several limitations to our work that we can improve upon. 
Our current implementation focuses on static scenes, and extending this approach to dynamic scenarios presents interesting challenges. 
Dynamic scenes would require not only temporal coherence in both Sketch and Patch Gaussian representations but also efficient updating mechanisms to handle moving objects and changing geometries. The effectiveness of our line segment-based encoding may vary in natural scenes where boundaries are less geometric, suggesting the exploration of alternative parametric models for organic shapes~\cite{yang2024linegs}. Additionally, scenes with complex, curved architectural features might benefit from more sophisticated geometric primitives beyond line segments.

Second, our method relies on hyperparameters tuning for balancing Sketch and Patch Gaussians. Future work could explore adaptive parameter selection based on scene characteristics and user requirements. The optimization of these parameters could potentially leverage machine learning techniques to automatically determine the optimal distribution between Sketch and Patch Gaussians based on scene geometry and desired quality-storage trade-offs.

Furthermore, while our approach significantly reduces storage requirements, it is complementary to existing progressive transmission strategies, such as the LOD approaches \cite{shi2025lapisgs, lu2024scaffold}. 
Our method's distinction between Sketch and Patch Gaussians naturally aligns with progressive streaming paradigms, where Sketch Gaussians could serve as an initial preview layer due to their structural importance, while Patch Gaussians could be transmitted progressively for quality refinement. 
This compatibility with layered representations opens possibilities for integrating our storage-efficient encoding with adaptive streaming systems, enabling both compact representation and flexible transmission based on available bandwidth and viewing requirements.

Another promising direction involves the integration of semantic understanding into the Gaussian categorization process. 
Currently, our method relies primarily on geometric features for categorization, but incorporating semantic information could lead to a more appropriate distribution of Gaussians based on object importance and viewer attention patterns. 
This could be particularly valuable in interactive applications where certain scene elements require higher visual fidelity than others.

\bibliographystyle{plain}
\bibliography{manuscript}

\begin{thebibliography}{10}

\bibitem{baillard1999automatic}
Caroline Baillard, Cordelia Schmid, Andrew Zisserman, and Andrew Fitzgibbon.
\newblock {Automatic line matching and 3D reconstruction of buildings from multiple views}.
\newblock In {\em ISPRS Conference on Automatic Extraction of GIS Objects from Digital Imagery}, volume~32, pages 69--80, 1999.

\bibitem{barron2022mip}
Jonathan~T Barron, Ben Mildenhall, Dor Verbin, Pratul~P Srinivasan, and Peter Hedman.
\newblock {{Mip-NeRF} 360: Unbounded anti-aliased neural radiance fields}.
\newblock In {\em Proceedings of the IEEE/CVF conference on computer vision and pattern recognition}, pages 5470--5479, 2022.

\bibitem{cai2025radiative}
Yuanhao Cai, Yixun Liang, Jiahao Wang, Angtian Wang, Yulun Zhang, Xiaokang Yang, Zongwei Zhou, and Alan Yuille.
\newblock {Radiative gaussian splatting for efficient x-ray novel view synthesis}.
\newblock In {\em European Conference on Computer Vision}, pages 283--299. Springer, 2025.

\bibitem{chen2017fast}
Siheng Chen, Dong Tian, Chen Feng, Anthony Vetro, and Jelena Kova{\v{c}}evi{\'c}.
\newblock {Fast resampling of three-dimensional point clouds via graphs}.
\newblock {\em IEEE Transactions on Signal Processing}, 66(3):666--681, 2017.

\bibitem{fan2023lightgaussian}
Zhiwen Fan, Kevin Wang, Kairun Wen, Zehao Zhu, Dejia Xu, and Zhangyang Wang.
\newblock {Lightgaussian: Unbounded 3D gaussian compression with 15x reduction and 200+ fps}.
\newblock {\em arXiv preprint arXiv:2311.17245}, 2023.

\bibitem{felzenszwalb2004efficient}
Pedro~F Felzenszwalb and Daniel~P Huttenlocher.
\newblock {Efficient graph-based image segmentation}.
\newblock {\em International journal of computer vision}, 59:167--181, 2004.

\bibitem{fischler1981random}
Martin~A Fischler and Robert~C Bolles.
\newblock {{Random sample consensus: a paradigm for model fitting with applications to image analysis and automated cartography}}.
\newblock {\em Communications of the ACM}, 24(6):381--395, 1981.

\bibitem{ipol.2012.gjmr-lsd}
Rafael Grompone~von Gioi, Jérémie Jakubowicz, Jean-Michel Morel, and Gregory Randall.
\newblock {LSD: a Line Segment Detector}.
\newblock {\em {Image Processing On Line}}, 2:35--55, 2012.
\newblock \url{https://doi.org/10.5201/ipol.2012.gjmr-lsd}.

\bibitem{han2020vivo}
Bo~Han, Yu~Liu, and Feng Qian.
\newblock {{ViVo}: Visibility-aware mobile volumetric video streaming}.
\newblock In {\em Proceedings of the 26th annual international conference on mobile computing and networking}, pages 1--13, 2020.

\bibitem{Hartley2000}
R.~I. Hartley and A.~Zisserman.
\newblock {\em {Multiple View Geometry in Computer Vision}}.
\newblock {Cambridge University Press, ISBN: 0521623049}, 2000.

\bibitem{hedman2018deep}
Peter Hedman, Julien Philip, True Price, Jan-Michael Frahm, George Drettakis, and Gabriel Brostow.
\newblock {Deep blending for free-viewpoint image-based rendering}.
\newblock {\em ACM Transactions on Graphics (ToG)}, 37(6):1--15, 2018.

\bibitem{hinton2015distilling}
Geoffrey Hinton.
\newblock {Distilling the Knowledge in a Neural Network}.
\newblock {\em arXiv preprint arXiv:1503.02531}, 2015.

\bibitem{hofer2014improving}
Manuel Hofer, Michael Maurer, and Horst Bischof.
\newblock {Improving sparse 3D models for man-made environments using line-based 3D reconstruction}.
\newblock In {\em 2014 2nd International Conference on 3D Vision}, volume~1, pages 535--542. IEEE, 2014.

\bibitem{hofer2015line3d}
Manuel Hofer, Michael Maurer, and Horst Bischof.
\newblock Line3d: Efficient 3d scene abstraction for the built environment.
\newblock In {\em Pattern Recognition: 37th German Conference, GCPR 2015, Aachen, Germany, October 7-10, 2015, Proceedings 37}, pages 237--248. Springer, 2015.

\bibitem{hofer2017efficient}
Manuel Hofer, Michael Maurer, and Horst Bischof.
\newblock {Efficient 3D scene abstraction using line segments}.
\newblock {\em Computer Vision and Image Understanding}, 157:167--178, 2017.

\bibitem{line3dcode}
Manuel Hofer, Michael Maurer, and Horst Bischof.
\newblock {Line3D++}.
\newblock {\scriptsize\url{https://github.com/manhofer/Line3Dpp}}, 2020.

\bibitem{huang2020tp}
Siyu Huang, Fangbo Qin, Pengfei Xiong, Ning Ding, Yijia He, and Xiao Liu.
\newblock {TP-LSD: Tri-points based line segment detector}.
\newblock In {\em European Conference on Computer Vision}, pages 770--785. Springer, 2020.

\bibitem{kahl2003multiview}
Fredrik Kahl and Jonas August.
\newblock {Multiview reconstruction of space curves}.
\newblock In {\em Proceedings Ninth IEEE International Conference on Computer Vision}, pages 1017--1024. IEEE, 2003.

\bibitem{kerbl20233D}
Bernhard Kerbl, Georgios Kopanas, Thomas Leimk{\"u}hler, and George Drettakis.
\newblock {{3D} {Gaussian} splatting for real-time radiance field rendering}.
\newblock {\em ACM Transactions on Graphics (ToG)}, 42(4):1--14, 2023.

\bibitem{gaussiancode}
Bernhard Kerbl, Georgios Kopanas, Thomas Leimk{\"u}hler, and George Drettakis.
\newblock {{3D Gaussian} Splatting for Real-Time Radiance Field Rendering}.
\newblock {\scriptsize\url{https://github.com/graphdeco-inria/gaussian-splatting}}, 2023.

\bibitem{kerbl2024hierarchical}
Bernhard Kerbl, Andreas Meuleman, Georgios Kopanas, Michael Wimmer, Alexandre Lanvin, and George Drettakis.
\newblock A hierarchical 3d gaussian representation for real-time rendering of very large datasets.
\newblock {\em ACM Transactions on Graphics (TOG)}, 43(4):1--15, 2024.

\bibitem{kim2024color}
Sieun Kim, Kyungjin Lee, and Youngki Lee.
\newblock {Color-cued Efficient Densification Method for 3D Gaussian Splatting}.
\newblock In {\em Proceedings of the IEEE/CVF Conference on Computer Vision and Pattern Recognition}, pages 775--783, 2024.

\bibitem{knapitsch2017tanks}
Arno Knapitsch, Jaesik Park, Qian-Yi Zhou, and Vladlen Koltun.
\newblock {Tanks and temples: Benchmarking large-scale scene reconstruction}.
\newblock {\em ACM Transactions on Graphics (ToG)}, 36(4):1--13, 2017.

\bibitem{lee2024compact}
Joo~Chan Lee, Daniel Rho, Xiangyu Sun, Jong~Hwan Ko, and Eunbyung Park.
\newblock {Compact 3D gaussian representation for radiance field}.
\newblock In {\em Proceedings of the IEEE/CVF Conference on Computer Vision and Pattern Recognition}, pages 21719--21728, 2024.

\bibitem{li20243d}
Lei Li, Songyou Peng, Zehao Yu, Shaohui Liu, R{\'e}mi Pautrat, Xiaochuan Yin, and Marc Pollefeys.
\newblock {3D Neural Edge Reconstruction}.
\newblock In {\em Proceedings of the IEEE/CVF Conference on Computer Vision and Pattern Recognition}, pages 21219--21229, 2024.

\bibitem{li2025geogaussian}
Yanyan Li, Chenyu Lyu, Yan Di, Guangyao Zhai, Gim~Hee Lee, and Federico Tombari.
\newblock {Geogaussian: Geometry-aware Gaussian splatting for scene rendering}.
\newblock In {\em European Conference on Computer Vision}, pages 441--457. Springer, 2025.

\bibitem{liu20233d}
Shaohui Liu, Yifan Yu, R{\'e}mi Pautrat, Marc Pollefeys, and Viktor Larsson.
\newblock {3D line mapping revisited}.
\newblock In {\em Proceedings of the IEEE/CVF Conference on Computer Vision and Pattern Recognition}, pages 21445--21455, 2023.

\bibitem{liu2021pc2wf}
Yujia Liu, Stefano D'Aronco, Konrad Schindler, and Jan~Dirk Wegner.
\newblock {PC2WF: 3D wireframe reconstruction from raw point clouds}.
\newblock {\em arXiv preprint arXiv:2103.02766}, 2021.

\bibitem{lu2024scaffold}
Tao Lu, Mulin Yu, Linning Xu, Yuanbo Xiangli, Limin Wang, Dahua Lin, and Bo~Dai.
\newblock {Scaffold-gs: Structured 3D gaussians for view-adaptive rendering}.
\newblock In {\em Proceedings of the IEEE/CVF Conference on Computer Vision and Pattern Recognition}, pages 20654--20664, 2024.

\bibitem{ma20223d}
Wenchao Ma, Bin Tan, Nan Xue, Tianfu Wu, Xianwei Zheng, and Gui-Song Xia.
\newblock {HoW-3D: Holistic 3D wireframe perception from a single image}.
\newblock In {\em 2022 International Conference on 3D Vision (3DV)}, pages 596--605. IEEE, 2022.

\bibitem{micusik2017structure}
Branislav Micusik and Horst Wildenauer.
\newblock {Structure from motion with line segments under relaxed endpoint constraints}.
\newblock {\em International Journal of Computer Vision}, 124:65--79, 2017.

\bibitem{mildenhall2021nerf}
Ben Mildenhall, Pratul~P. Srinivasan, Matthew Tancik, Jonathan~T. Barron, Ravi Ramamoorthi, and Ren Ng.
\newblock {NeRF}: representing scenes as neural radiance fields for view synthesis.
\newblock {\em Communications of the ACM}, 65(1):99–106, Dec 2021.

\bibitem{navaneet2023compact3d}
KL~Navaneet, Kossar~Pourahmadi Meibodi, Soroush~Abbasi Koohpayegani, and Hamed Pirsiavash.
\newblock {Compact3D: Compressing gaussian splat radiance field models with vector quantization}.
\newblock {\em arXiv preprint arXiv:2311.18159}, 2023.

\bibitem{niedermayr2024compressed}
Simon Niedermayr, Josef Stumpfegger, and R{\"u}diger Westermann.
\newblock {Compressed 3D gaussian splatting for accelerated novel view synthesis}.
\newblock In {\em Proceedings of the IEEE/CVF Conference on Computer Vision and Pattern Recognition}, pages 10349--10358, 2024.

\bibitem{papantonakis2024reducing}
Panagiotis Papantonakis, Georgios Kopanas, Bernhard Kerbl, Alexandre Lanvin, and George Drettakis.
\newblock Reducing the memory footprint of {3D Gaussian} splatting.
\newblock {\em Proceedings of the ACM on Computer Graphics and Interactive Techniques}, 7(1):1--17, 2024.

\bibitem{pautrat2023deeplsd}
R{\'e}mi Pautrat, Daniel Barath, Viktor Larsson, Martin~R Oswald, and Marc Pollefeys.
\newblock {DeepLSD: Line segment detection and refinement with deep image gradients}.
\newblock In {\em Proceedings of the IEEE/CVF Conference on Computer Vision and Pattern Recognition}, pages 17327--17336, 2023.

\bibitem{ramalingam2015line}
Srikumar Ramalingam, Michel Antunes, Dan Snow, Gim Hee~Lee, and Sudeep Pillai.
\newblock {Line-sweep: Cross-ratio for wide-baseline matching and 3D reconstruction}.
\newblock In {\em Proceedings of the IEEE Conference on Computer Vision and Pattern Recognition}, pages 1238--1246, 2015.

\bibitem{ren2024octree}
Kerui Ren, Lihan Jiang, Tao Lu, Mulin Yu, Linning Xu, Zhangkai Ni, and Bo~Dai.
\newblock {Octree-gs: Towards consistent real-time rendering with lod-structured 3D gaussians}.
\newblock {\em arXiv preprint arXiv:2403.17898}, 2024.

\bibitem{schindler2006line}
Grant Schindler, Panchapagesan Krishnamurthy, and Frank Dellaert.
\newblock {Line-based structure from motion for urban environments}.
\newblock In {\em Third International Symposium on 3D Data Processing, Visualization, and Transmission (3DPVT'06)}, pages 846--853. IEEE, 2006.

\bibitem{schmid2000geometry}
Cordelia Schmid and Andrew Zisserman.
\newblock {The geometry and matching of lines and curves over multiple views}.
\newblock {\em International Journal of Computer Vision}, 40:199--233, 2000.

\bibitem{schnabel2007efficient}
Ruwen Schnabel, Roland Wahl, and Reinhard Klein.
\newblock {Efficient RANSAC for point-cloud shape detection}.
\newblock In {\em Computer graphics forum}, volume~26, pages 214--226. Wiley Online Library, 2007.

\bibitem{schonberger2016structure}
Johannes~L Schonberger and Jan-Michael Frahm.
\newblock {Structure-from-Motion revisited}.
\newblock In {\em Proceedings of the IEEE conference on computer vision and pattern recognition}, pages 4104--4113, 2016.

\bibitem{shi2024qv4}
Yuang Shi, Bennett Clement, and Wei~Tsang Ooi.
\newblock {{QV4}: {QoE}-based Viewpoint-Aware {V-PCC}-encoded Volumetric Video Streaming}.
\newblock In {\em Proceedings of the 15th Conference on ACM Multimedia Systems}, 2024.

\bibitem{shi2025lapisgs}
Yuang Shi, Simone Gasparini, G{\'e}raldine Morin, and Wei~Tsang Ooi.
\newblock {LapisGS: Layered Progressive 3D Gaussian Splatting for Adaptive Streaming}.
\newblock In {\em The 12th International Conference on 3D Vision.}, 2025.

\bibitem{shi2023enabling}
Yuang Shi, Pranav Venkatram, Yifan Ding, and Wei~Tsang Ooi.
\newblock {Enabling Low Bit-Rate {MPEG} {V-PCC}-encoded Volumetric Video Streaming with {3D} Sub-sampling}.
\newblock In {\em Proceedings of the 14th Conference on ACM Multimedia Systems}, pages 108--118, 2023.

\bibitem{shi2024volumetric}
Yuang Shi, Ruoyu Zhao, Simone Gasparini, G{\'e}raldine Morin, and Wei~Tsang Ooi.
\newblock {Volumetric Video Compression Through Neural-based Representation}.
\newblock In {\em Proceedings of the 16th International Workshop on Immersive Mixed and Virtual Environment Systems}, pages 85--91, 2024.

\bibitem{viola2023volumetric}
Irene Viola and Pablo Cesar.
\newblock {Volumetric video streaming: Current approaches and implementations}.
\newblock {\em Immersive Video Technologies}, pages 425--443, 2023.

\bibitem{von2008lsd}
Rafael~Grompone Von~Gioi, Jeremie Jakubowicz, Jean-Michel Morel, and Gregory Randall.
\newblock {LSD: A fast line segment detector with a false detection control}.
\newblock {\em IEEE transactions on pattern analysis and machine intelligence}, 32(4):722--732, 2008.

\bibitem{wang2020pie}
Xiaogang Wang, Yuelang Xu, Kai Xu, Andrea Tagliasacchi, Bin Zhou, Ali Mahdavi-Amiri, and Hao Zhang.
\newblock {Pie-net: Parametric inference of point cloud edges}.
\newblock {\em Advances in neural information processing systems}, 33:20167--20178, 2020.

\bibitem{wang2004image}
Zhou Wang, Alan~C Bovik, Hamid~R Sheikh, and Eero~P Simoncelli.
\newblock {Image quality assessment: from error visibility to structural similarity}.
\newblock {\em IEEE transactions on image processing}, 13(4):600--612, 2004.

\bibitem{wei2022elsr}
Dong Wei, Yi~Wan, Yongjun Zhang, Xinyi Liu, Bin Zhang, and Xiqi Wang.
\newblock {{ELSR}: Efficient line segment reconstruction with planes and points guidance}.
\newblock In {\em Proceedings of the IEEE/CVF Conference on Computer Vision and Pattern Recognition}, pages 15807--15815, 2022.

\bibitem{wu2013towards}
Changchang Wu.
\newblock Towards linear-time incremental structure from motion.
\newblock In {\em 2013 International Conference on 3D Vision-3DV 2013}, pages 127--134. IEEE, 2013.

\bibitem{xue2024neat}
Nan Xue, Bin Tan, Yuxi Xiao, Liang Dong, Gui-Song Xia, Tianfu Wu, and Yujun Shen.
\newblock {NEAT: Distilling 3D Wireframes from Neural Attraction Fields}.
\newblock In {\em Proceedings of the IEEE/CVF Conference on Computer Vision and Pattern Recognition}, pages 19968--19977, 2024.

\bibitem{yang2024linegs}
Chenggang Yang, Yuang Shi, and Wei~Tsang Ooi.
\newblock Linegs: 3d line segment representation on 3d gaussian splatting.
\newblock {\em arXiv preprint arXiv:2412.00477}, 2024.

\bibitem{ye2023nef}
Yunfan Ye, Renjiao Yi, Zhirui Gao, Chenyang Zhu, Zhiping Cai, and Kai Xu.
\newblock {NEF: Neural edge fields for 3D parametric curve reconstruction from multi-view images}.
\newblock In {\em Proceedings of the IEEE/CVF Conference on Computer Vision and Pattern Recognition}, pages 8486--8495, 2023.

\bibitem{zhang2021elsd}
Haotian Zhang, Yicheng Luo, Fangbo Qin, Yijia He, and Xiao Liu.
\newblock Elsd: Efficient line segment detector and descriptor.
\newblock In {\em Proceedings of the IEEE/CVF International Conference on Computer Vision}, pages 2969--2978, 2021.

\bibitem{zhang2014structure}
Lilian Zhang and Reinhard Koch.
\newblock {Structure and motion from line correspondences: Representation, projection, initialization and sparse bundle adjustment}.
\newblock {\em Journal of Visual Communication and Image Representation}, 25(5):904--915, 2014.

\bibitem{zhang2018unreasonable}
Richard Zhang, Phillip Isola, Alexei~A Efros, Eli Shechtman, and Oliver Wang.
\newblock The unreasonable effectiveness of deep features as a perceptual metric.
\newblock In {\em Proceedings of the IEEE conference on computer vision and pattern recognition}, pages 586--595, 2018.

\bibitem{zwicker2001ewa}
Matthias Zwicker, Hanspeter Pfister, Jeroen Van~Baar, and Markus Gross.
\newblock {EWA volume splatting}.
\newblock In {\em Proceedings Visualization, 2001. VIS'01.}, pages 29--538. IEEE, 2001.

\end{thebibliography}

\end{document}